\documentclass[sigconf,nonacm]{acmart}

\AtBeginDocument{%
  }

\usepackage{balance}

\usepackage{caption}
\usepackage{subcaption}
\usepackage{colortbl}
\usepackage{multirow}
\usepackage{adjustbox}

\usepackage{tikz}
\definecolor{seed}{HTML}{FFF2CC}
\definecolor{seedborder}{HTML}{D6B554}
\definecolor{green}{HTML}{D5E8D4}
\definecolor{greenborder}{HTML}{7EB161}
\definecolor{red}{HTML}{F8CECC}
\definecolor{redborder}{HTML}{B7524E}
\definecolor{purple}{HTML}{E1D5E7}
\definecolor{purpleborder}{HTML}{906BA2}
\definecolor{grey}{HTML}{F5F5F5}
\definecolor{greyborder}{HTML}{666666}
\definecolor{blue2}{HTML}{DAE8FC}
\definecolor{blueborder}{HTML}{6689BD}
\definecolor{orange}{HTML}{FFE6CC}
\definecolor{orangeborder}{HTML}{D59700}
\definecolor{darkgrey}{HTML}{C0BCBC}

\begin{document}

\title[Relevant Entity Selection: KG Bootstrapping via Zero-Shot Analogical Pruning]{Relevant Entity Selection: Knowledge Graph Bootstrapping via Zero-Shot Analogical Pruning}

\author{Lucas Jarnac}
\orcid{0000-0002-2819-2679}
\affiliation{%
  \institution{Orange}
  \country{France}
}
\affiliation{%
  \institution{Université de Lorraine, CNRS, LORIA}
  \city{Nancy}
  \country{France}
}
\email{lucas.jarnac@orange.com}

\author{Miguel Couceiro}
\orcid{0000-0003-2316-7623}
\affiliation{%
  \institution{Université de Lorraine, CNRS, LORIA}
  \city{Nancy}
  \country{France}
}
\email{miguel.couceiro@loria.fr}

\author{Pierre Monnin}
\orcid{0000-0002-2017-8426}
\affiliation{%
  \institution{Orange}
  \country{France}
}
\email{pierre.monnin@orange.com}

\renewcommand{\shortauthors}{Lucas Jarnac, Miguel Couceiro, and Pierre Monnin}

\begin{abstract}
  Knowledge Graph Construction (KGC) can be seen as an iterative process starting from a high quality nucleus that is refined by knowledge extraction approaches in a virtuous loop. 
  Such a nucleus can be obtained from knowledge existing in an open KG like Wikidata. 
  However, due to the size of such generic KGs, integrating them as a whole may entail irrelevant content and scalability issues.
  We propose an analogy-based approach that starts from seed entities of interest in a generic KG, and keeps or prunes their neighboring entities.
  We evaluate our approach on Wikidata through two manually labeled datasets that contain either domain-homogeneous or -heterogeneous seed entities. 
  We empirically show that our analogy-based approach outperforms LSTM, Random Forest, SVM, and MLP, with a drastically lower number of parameters. 
  We also evaluate its generalization potential in a transfer learning setting.
  These results advocate for the further integration of analogy-based inference in tasks related to the KG lifecycle.
\end{abstract}



\maketitle

\section{Introduction}

Knowledge graphs (KGs) are ``graphs of data intended to accumulate and convey knowledge of the real world, whose nodes represent entities of interest and whose edges represent relations between these entities''~\cite{hoganBCAMGKENNNPRRSSSZ21}. 
More formally, KGs are directed and labeled multigraphs $(\mathcal{E}, \mathcal{R}, \mathcal{T})$, where
$\mathcal{E}$ is the set of entities, $\mathcal{R}$ is the set of relations, and $\mathcal{T}$ is the set of triples $\langle h, r, t \rangle \in \mathcal{E} \times \mathcal{R} \times \mathcal{E}$, where $r$ qualifies the relation holding between $h$ and $t$.
An example of such a triple could be $\langle \texttt{BarackObama}, \texttt{instanceOf}, \texttt{Person} \rangle$. 
KGs have proven useful in many academic and industrial applications, including search enhancement, question-answering, recommender systems, and eXplainable Artificial Intelligence~\cite{hoganBCAMGKENNNPRRSSSZ21,noyGJNPT19,tiddiS22}. 

\begin{figure}
    \centering
    \includegraphics[width=\linewidth]{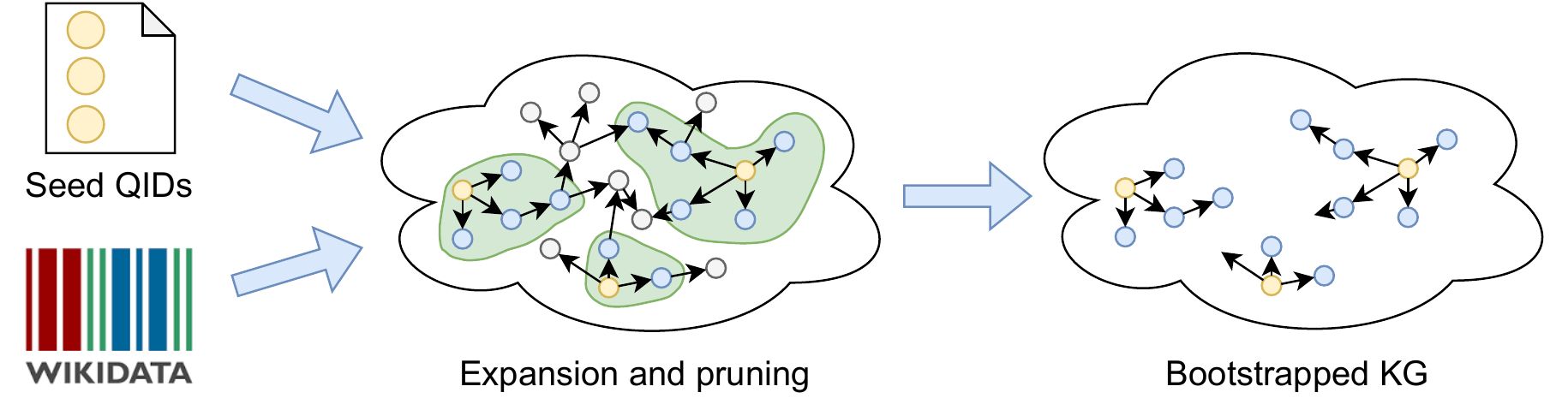}
    \caption{Outline of the bootstrapping of a knowledge graph from Wikidata.}
    \label{fig:bootstrapping-outline}
\end{figure}

Building and completing a KG can be achieved with knowledge extraction approaches from structured or unstructured data (\textit{e.g.}, tables, texts)~\cite{sequedaL2021,weikumDRS21}. 
This forms a virtuous loop in which the KG is both a supporting structure that provides entities and relations of interest to detect in data and the target structure to refine and complete. 
However, the cold start problem appears when the initial KG is empty, which motivates the need to build first a high quality nucleus~\cite{weikumDRS21}.
Such a nucleus could be manually bootstrapped by experts, but this process is time-consuming. 
Some authors thus propose to focus on premium sources of entities and categories to automatically constitute the nucleus~\cite{weikumDRS21}.
In this view, several works consider Wikidata~\cite{vrandecic}, a large and generic KG collaboratively built to support Wikipedia, as a premium source~\cite{jarnacM22,shbitaGLDR23}.
However, the sheer size of Wikidata entails a need to restrict Wikidata knowledge to be integrated into the KG nucleus to avoid irrelevant knowledge and scalability issues.
As Wikidata contains more than 100 million entities\footnote{\url{https://www.wikidata.org/wiki/Wikidata:Statistics}}, authors adopt a distillation~\cite{shbitaGLDR23} or a pruning~\cite{jarnacM22} process, in which seed entities\footnote{Note that entities in Wikidata are identified by QIDs.} of interest are identified in Wikidata and only parts of their neighborhood are included in the KG nucleus (see Figure~\ref{fig:bootstrapping-outline}).
The selection of the neighboring entities leverages the ontology hierarchy, either only upward~\cite{shbitaGLDR23}, or both upward and downward~\cite{jarnacM22}.
The latter brings much more entities, which makes it more prone to gathering irrelevant knowledge in the KG nucleus.
For instance, the downward neighbors of \texttt{Microsoft SharePoint} include  \texttt{Content Management System}, a relevant entity to keep, and \texttt{Dating App}, an irrelevant one to prune.
Jarnac and Monnin~\cite{jarnacM22} thus use several pruning thresholds based on node degrees and distances in the embedding space but highlight the difficulty to set global thresholds when applied to heterogeneous entities with different distributions of degrees and distances.
Additionally, to the best of our knowledge, there is no publicly available benchmark dataset to evaluate such approaches.

In our work, we propose to tackle the limitations of fixed thresholds by training classifiers to select (or keep) relevant neighboring entities and prune irrelevant ones in a KG bootstrapping process. 
Specifically, we propose an analogy-based zero-shot approach.
Analogies are quadruples of the form $\texttt{Paris} : \texttt{France} :: \texttt{Berlin} : \texttt{Germany}$, which can be read ``Paris is to France as Berlin is to Germany'', that simultaneously capture similarities and dissimilarities between objects~\cite{miclet2008,mitchell21}. 
Analogical reasoning is a remarkable capability of the human mind, that has recently obtained impressive results on NLP tasks when applied on character and word embeddings~\cite{lim19,marquer22,sultanS22}. 
Such a reasoning has also been proposed for KGs~\cite{ilievskiPS22,liuWY17,monninC2022,portischHP22,yaoZCHZC2023}, using KG embeddings, \textit{i.e.}, vector representations of KG entities and relations that preserve as much as possible the properties of the graph~\cite{caiZC18}. 

In our approach, we combine analogical reasoning and graph embedding to keep or prune neighboring entities.
Our intuition is that the analogy-based model will be able to capture relative similarities and dissimilarities between seed entities and their neighbors to keep or to prune, and thus avoid the caveats of fixed thresholds that have difficulty generalizing to heterogeneous entities.
Furthermore, our approach is zero-shot: the model learns to detect relative similarities and dissimilarities on a set of seed entities and their neighbors and can extrapolate to unseen seed entities and their neighbors.
We experiment our approach on the Wikidata KG and two annotated datasets of seed entities and their neighbors to keep or prune, and that we make available for the benefit of the community.
We empirically compare the behavior of our approach with several classifiers (\textit{e.g.}, Random Forest, LSTM) and symbolic approaches (\textit{e.g.}, depth pruning, threshold pruning~\cite{jarnacM22}). 
We also assess the performance of the different approaches based on evaluation metrics and number of parameters. Moreover, we test the generalization of this analogy-based approach on a transfer learning setting. 

The main contributions of the paper are:
\begin{itemize}
    \item We propose an analogy-based zero-shot approach to select relevant entities in the neighborhood of seed entities. This approach only needs training examples of entities to keep or prune for some seed entities and can extrapolate to new seed entities without selection / pruning examples for them. 
    \item We present a comparative study of our analogy-based  approach to other methodologies with respect to different performance metrics, the number of parameters to be trained and the generality of the models considered.    
    \item We provide two annotated datasets of seed entities and relevant or irrelevant neighbors in Wikidata to start constituting publicly available benchmarks for the community.
\end{itemize}

The remainder of this article is structured as follows. 
We briefly survey related work about KG bootstrapping and analogy-based inference in Section~\ref{section:related-work}, and we  detail our analogy-based zero-shot approach to select relevant entities to bootstrap a KG  in
Section~\ref{section:analogical-pruning}. 
We experiment with Wikidata and two datasets in Section~\ref{section:experiments} and we discuss our results in Section~\ref{section:discussion}.
Finally, Section~\ref{section:conclusion} summarizes our work and outlines future research work.

\section{Related work}
\label{section:related-work}

Our work positions within approaches focusing on bootstrapping KGs, especially by pruning to limit the scope of the built KG. 
We review some prominent works of this line of research in Subsection~\ref{subsection:kg-bootstrapping}.
Additionally, we rely on analogical reasoning which has recently achieved significant performance on NLP-related tasks, and has been identified as a promising research direction for KG-related tasks as outlined in Subsection~\ref{subsection:analogical-reasoning}.

\subsection{KG Bootstrapping and Pruning}
\label{subsection:kg-bootstrapping}

The construction of ontologies and KGs usually entails the possibility of their reuse for other purposes.
That is why, it is common to leverage existing ontologies and KGs to bootstrap others~\cite{fernandez1997, swartoutPKR1996}, since they can be seen as premium sources~\cite{weikumDRS21}.
To illustrate, YAGO3 combines the taxonomy of WordNet and the categories of Wikipedia pages~\cite{mahdisoltaniBS15}, Knowledge Vault integrates the FreeBase KG~\cite{dongGHHLMSSZ14}, and PGxLOD first integrates several biomedical KGs to then interconnect and enrich them~\cite{monninLHRTJNC19}.

Due to the size and generic aspect of some KGs and ontologies, some authors resort to pruning to construct domain-specific KGs from them.
One of the early examples is the work of Swartout \textit{et al.} in 1996~\cite{swartoutPKR1996} where they propose to build a domain-specific ontology starting from a large and generic ontology, SENSUS, of 50,000+ concepts.
To do so, they start with seed concepts from the domain of interest that are manually linked to SENSUS.
They then include all super-concepts up to the root.
They also discuss that some subtrees bring additional concepts of interest.
They manually identify them with the rationale that if some nodes of a subtree have been identified relevant, then the other nodes of the subtree may be relevant too.
However, such a manual process is time-consuming.
Furthermore, incorporating subtrees of large ontologies may come at the expense of incorporating irrelevant knowledge, which is difficult to manually assess. 
To face such issues, automatic distillation or pruning approaches can be considered.
The distillation process can be guided by documents from the domain of interest.
For instance, Babayeva \textit{et al.} develop a domain ontology for Cyber Defence exercises by collecting concepts from an existing ontology and documents on this topic~\cite{babayeva2022}.
Shbita \textit{et al.}~\cite{shbitaGLDR23} build a KG about customer requirements starting from client verbatim and Wikidata. 
Specifically, they detect entities in text, link them to Wikidata, and integrate their direct classes and all their super-classes.

Regarding pruning approaches, they can be classified into two categories: \textit{aggressive pruning} based on topology of the graph and \textit{soft pruning} that requires human input to define the relevant taxonomic concepts.
In~\cite{faralli18}, Faralli \textit{et al.} introduce the CrumbTrail algorithm that prunes a directed noisy knowledge graph with the aim of obtaining an acyclic subgraph that contains all previously selected seed nodes.
Using this CrumbTrail algorithm, Bordea \textit{et al.} propose to build domain-specific taxonomies from the KG of Wikipedia categories.
After a user has selected leaf nodes and a root node, the algorithm is applied on the KG to build the directed and acyclic graph that will form the taxonomy. 
They provide three datasets but they are not directly applicable to our task. 
Indeed, they have specific concerns w.r.t. to taxonomy building (\textit{e.g.}, upward extension from leaves to root, acyclic aspect) while we mainly focus on gathering terms of interest w.r.t. a domain without such concerns.
Additionally, we consider that seed nodes may not be leaves and propose to perform a downward expansion (see Subsection~\ref{subsection:expansion}).
This is a more difficult pruning task since not all subclasses of a class of interest may be relevant. 
In this view, Jarnac and Monnin~\cite{jarnacM22} bootstrap an enterprise KG by expanding a set of business terms semi-manually aligned to Wikidata entities along their ontology hierarchy.
According to the authors, the distance in the embedding space appears to be a good indicator of topic similarity or drift.
Thus, to limit the expansion, they propose an automatic approach relying on node degree and distance thresholds.
However, such thresholds are globally fixed for all seed entities, which may lead to varying performance when these entities belong to heterogeneous domains.

\subsection{Analogy-Based Inference in KGs}
\label{subsection:analogical-reasoning}

Analogy-based inference is a basic process in human cognition~\cite{mitchell21,Chollet} that is tightly related to  abstraction, adaptation and creativity.  Analogy-based inference can be viewed as transferring knowledge from a source domain to a different, but somewhat
similar, target domain by leveraging simultaneously on similarities and dissimilarities. 
Most of the literature in analogy-based inference is built on the notion of \emph{analogical proportions}, {\it i.e.,}  statements of the form ``$A$ is to $B$ as $C$ is to $D$''
represented as $A : B :: C : D$~\cite{miclet2008}, and relies on two main tasks, namely, {\it analogy detection} that involves deciding whether a quadruple $(A, B, C, D)$ constitutes a valid analogy $A : B :: C : D$, and {\it analogy solving} that consists in finding the possible elements $X$ that make $A : B :: C : X$ a valid analogy. 

When the underlying objects $A,B,C$ and $D$ are represented as vectors $e_{A},e_{B},e_{C}$ and $e_{D}$, respectively, in some vector space $\mathbb{R}^n$, analogical proportions can be thought of in geometric terms as the parallelogram rule $e_{D} - e_{C} = e_{B} - e_{A}$. For instance, the underlying elements $A,B,C$ and $D$ of the analogical proportion could be words~\cite{Turney.08}, and $e_{A},e_{B},e_{C}$ and $e_{D}$ their vectorial representations \cite{MikolovNIPS2013,efficient-representation-w2v:2013:mikolov} or larger chunks of text such as sentences \cite{zhu-de-melo-2020-sentence,afantenos.2021,afantenos.2022}.
Analogy-based  inference has been used to solve hard reasoning tasks and has shown its potential with competitive
results in several  machine learning tasks such as classification, decision making and recommendation \cite{FahandarH18,FahandarH21,HugPRS19,galois-analogical-clf:2023:couceiro-lehtonen}, in  data augmentation through analogical  extrapolation for model
learning, especially in environments with few labeled examples \cite{extend-boolean:2017:couceiro,boolean-analogy:2018:couceiro}. Moreover, it has been successfully applied in classical  natural language processing
(NLP) tasks such as machine translation \cite{analogy-alea:2009:langlais}, several semantic ~\cite{lim19,analogies-ml:2021:lim} and morphological tasks~\cite{minimal-complexity:2020:murena,detecting:2021:alsaidi,marquer22}, as well as in  (visual) question answering ~\cite{SadeghiZF15}, solving puzzles and
scholastic aptitude tests  \cite{PeyreSLS19}, and target sense verification \cite{Zervakis}.

Analogy-based inference can also be used to address and tackle tasks related to the KG lifecycle such as semantic table interpretation or knowledge matching~\cite{monninC2022}.
A few works already exist in this line of research and rely on graph embedding, similarly to NLP approaches relying on character or word embeddings. 
For example, Liu \textit{et al.}~\cite{liuWY17} tackles the task of link prediction, \textit{i.e.}, completing triples $(h, r, ?)$, and study whether KG embedding models respect the parallelogram rule of analogical inference. 
They show that modeling analogical structures in KG embedding models brings additional performance.
Similarly, Yao \textit{et al.}~\cite{yaoZCHZC2023} propose a model based on analogy functions to enhance a KG embedding model for link prediction.
Alternatively, Portisch \textit{et al.}~\cite{portischHP22} evaluate whether KG embedding models for link prediction or data mining can be used for analogy detection.
In a similar fashion, we propose to leverage KG embeddings in an analogy detection task, where analogies serve to detect relevant or irrelevant entities w.r.t. seed entities of interest.

\section{Analogy-based zero-shot selection of relevant entities}
\label{section:analogical-pruning}

We consider that we have at our disposal a set of seed entities of interest.
Such entities can be identified (semi-)manually by experts and/or via an automatic extraction of entities from texts in the domains of interest~\cite{jarnacM22,shbitaGLDR23,swartoutPKR1996}. 
These seed entities are aligned with a generic KG, \textit{e.g.}, Wikidata, and we retrieve their neighboring entities of interest along the ontology hierarchy.
We first describe how we traverse the ontology hierarchy (Subsection~\ref{subsection:expansion}), and then how we keep relevant entities and prune irrelevant ones during this traversal using an analogy-based model (Subsection~\ref{subsection:analogy-model}).

\subsection{Expansion Along the Ontology Hierarchy}
\label{subsection:expansion}

\begin{figure}
    \centering
    \includegraphics[width=0.9\linewidth]{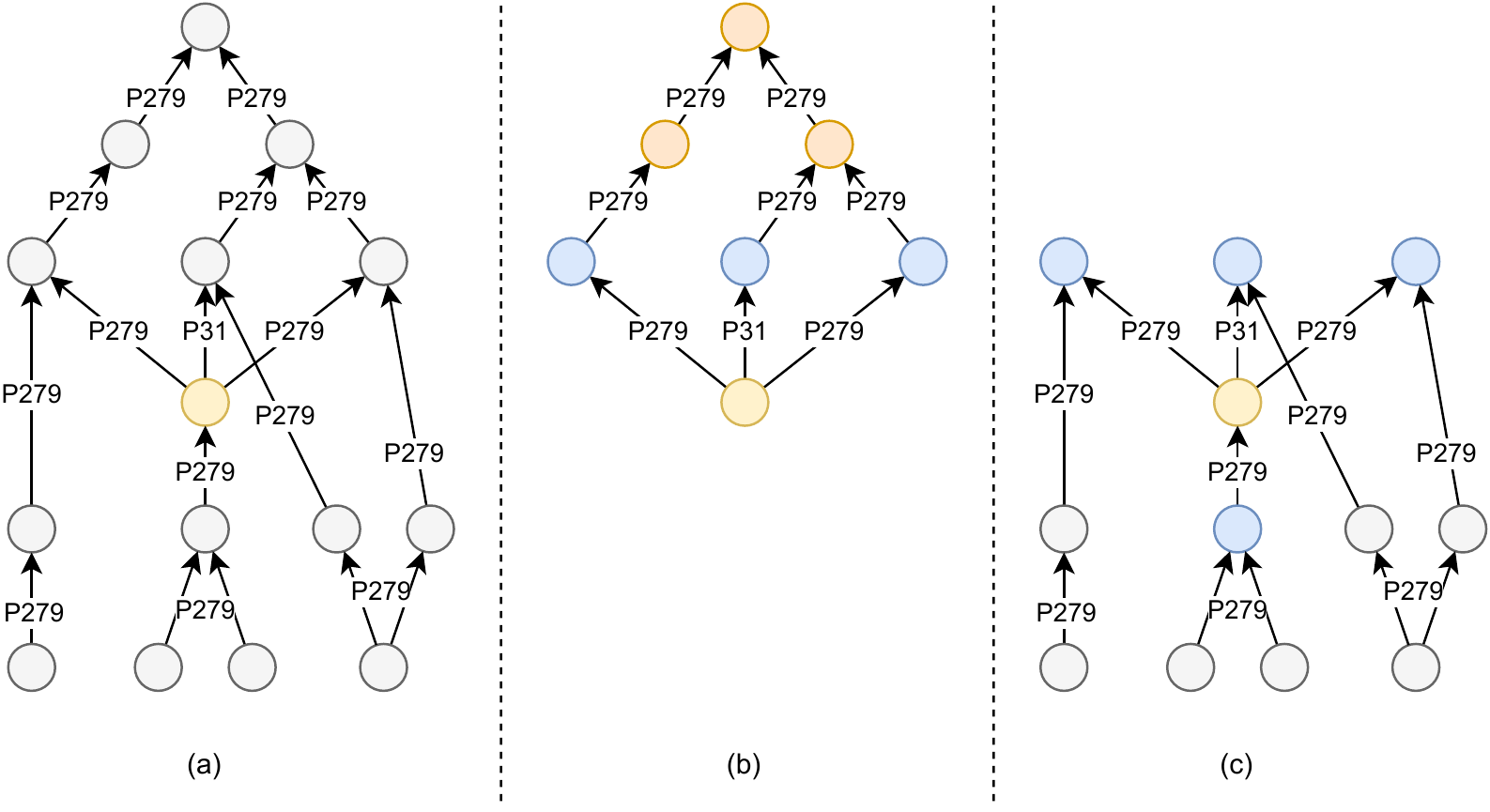}
    \caption[]{Expansion from a seed entity \raisebox{-1pt}{\begin{tikzpicture}\node[draw=seedborder, circle, fill=seed]{} ;\end{tikzpicture}} along the ontology hierarchy. P31 stands for ``instance of'' and P279 stands for ``subclass of''. 
    (a) depicts the full ontology hierarchy, (b) depicts classes reached in the upward expansion, and (c) depicts classes reached in the downward expansion.}
    \label{fig:hierarchical-expansion}
\end{figure}
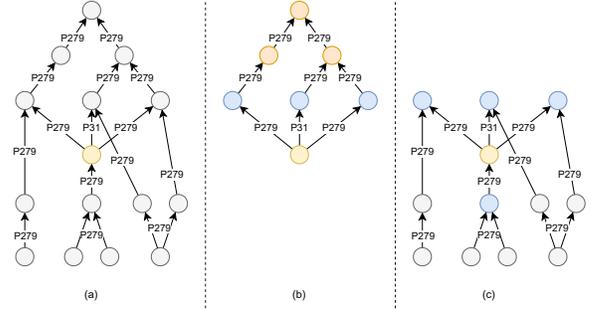

We retrieve the neighboring entities of seed entities of interest along the ontology hierarchy as illustrated in Figure~\ref{fig:hierarchical-expansion}, which distinguishes two directions for the expansion.

In the upward expansion (Figure~\ref{fig:hierarchical-expansion}b),  we retrieve from a seed entity \raisebox{-1pt}{\begin{tikzpicture}\node[draw=seedborder, circle, fill=seed]{} ;\end{tikzpicture}}, its first-level classes \raisebox{-1pt}{\begin{tikzpicture}\node[draw=blueborder, circle, fill=blue2]{} ;\end{tikzpicture}} following P31 (``instance of'') and P279 (``subclass of'') edges, \textit{i.e.}, we retrieve the classes that the entity directly instantiates, or those that directly subsume it. 
Then, we retrieve all their superclasses \raisebox{-1pt}{\begin{tikzpicture}\node[draw=orangeborder, circle, fill=orange]{} ;\end{tikzpicture}} by following P279 edges up to the root of the hierarchy in a breadth-first search expansion. 

In the downward expansion (Figure~\ref{fig:hierarchical-expansion}c),  we retrieve from a seed entity \raisebox{-1pt}{\begin{tikzpicture}\node[draw=seedborder, circle, fill=seed]{} ;\end{tikzpicture}}, its first-level classes \raisebox{-1pt}{\begin{tikzpicture}\node[draw=blueborder, circle, fill=blue2]{} ;\end{tikzpicture}} by following P31, P279, and reversed P279 edges, \textit{i.e.}, we retrieve the classes that the entity directly instantiates, those that directly subsume it, or those that it directly subsumes.
We then retrieve all their subclasses \raisebox{-1pt}{\begin{tikzpicture}\node[draw=greyborder, circle, fill=grey]{} ;\end{tikzpicture}} following reversed P279 edges up to the leaves, in a breadth-first search expansion.

\subsection{Selection of Relevant Entities}
\label{subsection:analogy-model}

In this subsection, we focus on the problem of keeping relevant and pruning irrelevant entities when traversed during the expansion described in Subsection~\ref{subsection:expansion}. 

\subsubsection{Formalization}
\label{subsubsection:formalization}

We formalize the problem as follows: given a seed entity $e_s$ and an entity $e_r$ reached during the graph expansion from $e_s$, our goal is to decide whether to keep or to prune $e_r$. 
If $e_r$ is kept, then its neighbors are explored as described in Subsection~\ref{subsection:expansion}.
Otherwise, its neighbors are not explored.

We propose an analogy-based zero-shot classifier model $\mathcal{A}$ such that:
\begin{equation}
    \mathcal{A}(e_s, e_r) = \begin{cases}
    1 \text{ if $e_r$ should be kept} \\
    0 \text{ if $e_r$ should be pruned}
    \end{cases}
\end{equation}
Recall that analogies are statements of the form ``A is to B as C is to D'' represented as $\text{A : B :: C : D}$~\cite{marquer22}.
In our case, we use analogies of the form 
\begin{equation}
    e_s^1 : e_r^1 :: e_s^2 : e_r^2
    \label{eq:analogy}
\end{equation}
where $e_s^1$ and $e_s^2$ are two seed entities, $e_r^1$ is reached during the expansion from $e_s^1$, and $e_r^2$ is reached during the expansion from $e_s^2$. 
Using analogical inference, if Equation~\eqref{eq:analogy} is a valid analogy and we know the decision for the pair $(e_s^1, e_r^1)$, then we can extrapolate the decision for the pair $(e_s^2, e_r^2)$.
To illustrate, if the analogy $\texttt{Hadoop} : \texttt{Big Data} :: \texttt{SharePoint} : \texttt{Content Management System}$ is valid and we know that Big Data should be kept, then Content Management System should also be kept.

We propose three different configurations for our analogy-based classifier, that consider different valid and invalid analogies.
To ease notation, we note k a keeping decision for a pair, and p a pruning decision for a pair. 
The three configurations are as follows:
\begin{description}
    \item[Configuration $\mathcal{C}_1$] Valid analogies are of the form $k :: k$. Invalid analogies are of the form $k :: p$.
    \item[Configuration $\mathcal{C}_2$] Valid analogies are of the form $k :: k$. Invalid analogies are of the form $k :: p$ and $p :: p$.
    \item[Configuration $\mathcal{C}_3$] Valid analogies are of the form $k :: k$ and $p :: p$. Invalid analogies are of the form $p :: k$ and $k :: p$
\end{description}
It is worth noting that depending on the chosen configuration, the aforementioned analogical inference is adapted. 
For example, with $\mathcal{C}_1$ and $\mathcal{C}_2$, valid analogies only allow to extrapolate keeping decisions. 
On the contrary, with $\mathcal{C}_3$, valid analogies can conclude on keeping or pruning the pair $(e_s^2, e_r^2)$, depending on the known decision for the pair $(e_s^1, e_r^1)$.
The same rationale can be applied on invalid analogies.
For example, with $\mathcal{C}_1$, invalid analogies lead to a pruning decision for the pair $(e_s^2, e_r^2)$.
On the contrary, with $\mathcal{C}_3$, invalid analogies lead to decide for the pair $(e_s^2, e_r^2)$ the opposite decision of the pair $(e_s^1, e_r^1)$.

In addition to configurations, we propose to consider or not paths in the graph within the analogy-based model.
To illustrate, consider the analogy in Equation~\eqref{eq:analogy} and the two following expansion paths that generated it:
\begin{align*}
    & e_s^1 \xrightarrow{} e_r^3 \xrightarrow{} e_r^1 \\
    & e_s^2 \xrightarrow{} e_r^4 \xrightarrow{} e_r^5 \xrightarrow{} e_r^2
\end{align*}
Our first formalization in Equation~\eqref{eq:analogy} only considers seed entities and reached entities, on which a decision is known or a decision is to be made by the model.
We also propose to consider the paths leading to the reached entities. 
In this view, the analogy in Equation~\eqref{eq:analogy} becomes as follows:
\begin{equation}
    e_s^1 : (e_r^3, e_r^1) :: e_s^2 : (e_r^4, e_r^5, e_r^2).
    \label{eq:path-analogy}
\end{equation}

\subsubsection{Model} 
\label{subsubsection:model}

\begin{figure}
    \centering
    \includegraphics[width=0.9\linewidth]{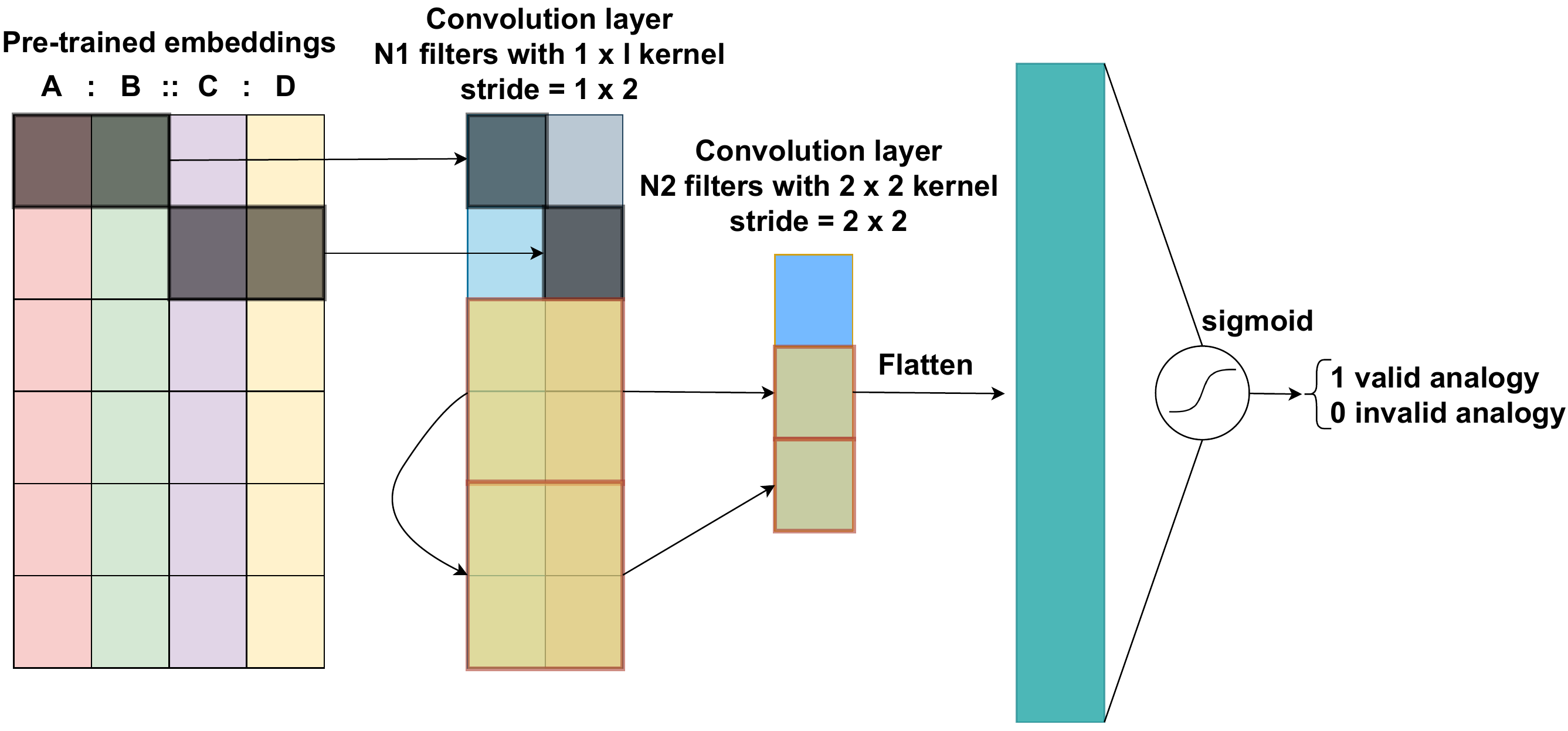}
    \caption{The analogy-based classifier model with two convolution layers and one fully connected layer from~\cite{lim19}.}
    \label{fig:analogy-model}
\end{figure}

We adopt the supervised machine learning model proposed by Lim \textit{et al.}~\cite{lim19}. 
This model is presented in Figure~\ref{fig:analogy-model} and relies on convolutional neural networks (CNNs).
It takes as input the vector embeddings of each constituent of a quadruple.
In our case, for a quadruple without paths (\textit{i.e.}, Equation~\eqref{eq:analogy}), we simply concatenate the embeddings of the entities (\textit{i.e.}, $e_s^1$, $e_r^1$, $e_s^2$, $e_r^2$). 
For a quadruple with paths (\textit{i.e.}, Equation~\eqref{eq:path-analogy}), we concatenate the embeddings of the entities (\textit{i.e.}, $e_s^1$, $e_r^3$, $e_r^1$, $e_s^2$, $e_r^4$, $e_r^5$, $e_r^2$) and use zero-padding in order to respect the model's fixed input dimension.
We experimented with three zero-padding methods:
\begin{description}
    \item[before] zeros are added before the sequence (\textit{e.g.}, before $e_s^1$ and before $e_s^2$)
    \item [between] zeros are added between the embedding of the seed entity and the embeddings of the entities in the path (\textit{e.g.}, between $e_s^1$ and $e_r^3$, and between $e_s^2$ and $e_r^4$)
    \item [after] zeros are added after the embeddings of the entities in the path (\textit{e.g.}, after $e_r^1$ and after $e_r^2$)
\end{description}

The model has two convolution layers, followed by a flattening operation and one fully connected layer. 
The first convolution layer is composed of $n_{1}$ filters. Each filter has a kernel size of $1 \times \text{sequence length}$ and a $\mathrm{ReLU}$ activation function. Filters are initialized with a He normal initializer and applied with a stride of $(1, \text{sequence length})$.
The second convolution layer is composed of $n_{2}$ filters. Each filter has a kernel size of $2 \times 2$ and a $\mathrm{ReLU}$ activation function. Filters are initialized with a He normal initializer and applied with a stride of $(2, 2)$. 
The last layer is a fully connected layer with one output and a sigmoid activation function to obtain a binary classification score in $\left[0, 1\right]$.
We also add dropout after each convolution layer.

This model is well-suited to the task at hand.
Indeed, the first convolution layer allows to compute dissimilarities for each pair of entities, while the second convolution layer compares these dissimilarities between the first and the second pairs forming the quadruple to classify as a valid or invalid analogy.

\subsubsection{Training}

We consider that we have at our disposal pairs $(e_s^1, e_r^1)$ whose keeping or pruning decision is known (\textit{e.g.}, annotations by experts).

To train our model, for each annotated pair $(e_s^1, e_r^1)$, and for each form of valid and invalid analogies of the considered configuration, we build $M$ analogies by sampling $M$ other adequate labeled pairs.
To illustrate, in configuration $\mathcal{C}_1$, invalid analogies are of the form $k :: p$. 
Thus, for a pair $(e_s^1, e_r^1)$ whose decision is keep, we build $M$ invalid analogies by selecting $M$ other pairs whose decision is prune.
These $M$ pairs are selected by ascending order of proximity of seed entities in the embedding space.
We then train our model by minimizing the binary cross-entropy loss and taking into account possible unbalancing between valid and invalid analogies.

\subsubsection{Inference}

At inference, on an unknown pair $(e_s^2, e_r^2)$:
\begin{enumerate}
    \item We select $N$ pairs $(e_s^1, e_r^1)$ whose decision is known to be keeping and $N$ pairs $(e_s^1, e_r^1)$ whose decision is known to be pruning.
    Specifically, for each type of decision, we order known pairs by ascending proximity of $e_s^1$ and $e_s^2$ in the embedding space and select the $N$ first. 
    \item We generate $2N$ quadruples with selected known pairs on the left and the unknown pair $(e_s^2, e_r^2)$ on the right\footnote{It is noteworthy that for configuration $\mathcal{C}_1$, the $N$ pairs $(e_s^1, e_r^1)$ whose decision is pruning are not used, leading to only $N$ quadruples being generated.}.
    \item For each of these quadruples, our model predicts whether it is a valid or invalid analogy, which constitutes a keeping or pruning prediction, depending on the chosen configuration (see Subsubsection~\ref{subsubsection:formalization}).
    \item We compute the average of the scores output by the model (in $\left[0, 1\right]$) on each of the $2N$ quadruples as follows:
    \begin{itemize}
        \item For $\mathcal{C}_1$: we interpret the score as a vote for keeping
        \item For $\mathcal{C}_2$: we interpret the score as a vote for keeping
        \item For $\mathcal{C}_3$: 
        \textit{(i)} if the known pair $(e_s^1, e_r^1)$ has a keeping decision, the score is considered as a vote for keeping; \textit{(ii)} if the known pair $(e_s^1, e_r^1)$ has a pruning decision, the score is considered as a vote for pruning. 
        Indeed, in this case, a score close to 1 corresponds to a valid analogy of the form $p :: p$.
        A score close to 0 corresponds to an invalid analogy of the form $p :: k$.
        Thus, we use $1 - score$ as a vote for keeping.
    \end{itemize}
    \item If the averaged keeping score is above a fixed threshold, we keep $e_r^2$. Otherwise we prune it.
\end{enumerate}
It should be noted that, at inference, our model extrapolates on pairs in which $e_s^2$, and potentially $e_r^2$ were not seen in training.
This makes our approach fundamentally zero-shot.

\section{Experiments}
\label{section:experiments}

We evaluate our analogy-based model on the Wikidata knowledge graph~\cite{vrandecic} and two datasets containing seed entities and labeled keeping and pruning decisions for their neighboring entities.
In particular, we compare the latter with baseline models such as Multi-Layer Perceptron (MLP), Long Short Term Memory (LSTM), Support Vector Machine (SVM), Random Forest, depth pruning, and threshold pruning~\cite{jarnacM22}.
We also evaluate our model in a transfer learning setting.

Since our approach requires KG embeddings, we experiment with the pre-trained embeddings of Wikidata available in PyTorch-BigGraph~\cite{lerer2019}\footnote{\url{https://torchbiggraph.readthedocs.io/en/latest/pretrained_embeddings.html}}.
These embeddings were learned for more than 78,000,000 entities of the 2019-03-06 version of Wikidata.
For building, training, and evaluating our models we used TensorFlow's Keras API and scikit-learn~\cite{scikit-learn}.
Datasets\footnote{\url{https://doi.org/10.5281/zenodo.8091584}} and code\footnote{\url{https://github.com/Orange-OpenSource/analogical-pruning}} of our experiments are publicly available.

\subsection{Datasets}
\label{subsection:datasets}

To the best of our knowledge, there is no publicly available benchmark dataset for the present task. 
This motivated us to build and publicly release the two datasets whose characteristics are detailed in Table~\ref{tab:datasets-statistics}.
Specifically, we gathered two sets of seed entities: 455 seed entities from the Computer Science / Information Technology domain for Dataset 1 (\textit{e.g.,} entities related to telecommunications, network, or programming languages), and 105 seed entities from more heterogeneous domains for Dataset 2 (\textit{e.g.}, entities related to food, music, sport, or science).
Table~\ref{tab:datasets-statistics} shows the number of nodes reached with an unconstrained (\textit{i.e.,} without pruning) upward and downward expansion as described in Section~\ref{subsection:expansion} for both datasets.
It can be noticed that the number of reached nodes upward is drastically lower than the number of reached nodes downward.
This motivated us to solely focus on pruning during the downward expansion.

To obtain labeled keeping and pruning decisions for downward nodes for both datasets without having to label the whole neighborhood, we adopted the following process.
We performed a downward expansion with the pruning approach proposed by Jarnac and Monnin~\cite{jarnacM22} with thresholds based on node degrees and distance in the embedding space. 
To configure these thresholds, we set $\alpha = 1.5$, $\gamma = 20$, $\beta = 1.3$ following~\cite{jarnacM22}.
Accordingly to their proposal, we also consider two different embeddings for entities: \textit{(i)}
Embedding $\mathcal{E}_{1}$ in which the embedding of an entity is its vector in the considered pre-trained embeddings, and \textit{(ii)} Embedding $\mathcal{E}_{2}$ in which the embedding of an entity is the centroid of the embeddings of its instances. If an entity does not have instances, its pre-trained embedding vector is used instead.
Then, we manually labeled keeping and pruning decisions output by this approach on the two sets of seed entities.
Note that, since we use pre-trained embeddings from 2019 and a Wikidata dump from 2022, some entities do not have embeddings.
To deal with this problem, we filtered both datasets to ensure that all seed and reached entities have an embedding vector.

\begin{table*}
    \centering
    \begin{tabular}{llrrrrrrr}
        \toprule
         \multicolumn{2}{c}{Dataset} & \multicolumn{1}{c}{\# Seed entities} & \multicolumn{1}{c}{\# Nodes up} & \multicolumn{1}{c}{\# Nodes down} & \multicolumn{1}{c}{\# P decisions} & \multicolumn{1}{c}{Depths P} & \multicolumn{1}{c}{\# K decisions} & \multicolumn{1}{c}{Depths K} \\
         \midrule
         \multirow{2}{*}{Dataset 1} & w/o filtering & 455 & 1,507 & 2,593,609 & 3,464 & $\llbracket 1, 4 \rrbracket$ & 1,769 & $\llbracket 1, 4 \rrbracket$ \\
         & w/ filtering & 439 & 1,469 & 2,593,575 & 2,910 & $\llbracket 1, 4 \rrbracket$ & 1,619 & $\llbracket 1, 4 \rrbracket$ \\
         \midrule
         \multirow{2}{*}{Dataset 2} & w/o filtering & 105 & 1,159 & 1,247,385 & 388 & $\llbracket 1, 2 \rrbracket$ & 594 & $\llbracket 1, 3 \rrbracket$\\ 
         & w/ filtering & 104 & 1,152 & 1,247,383 & 314 & $\llbracket 1, 2 \rrbracket$ & 577 & $\llbracket 1, 3 \rrbracket$\\ 
        \bottomrule
    \end{tabular}
    \caption{Statistics of Dataset 1 and Dataset 2 before and after filering to only retain entities with embeddings. K stands for Keeping and P stands for Pruning.}
    \label{tab:datasets-statistics}
\end{table*}

\subsection{Experimental Setup}
\label{subsection:experimental-setup}
 
We now describe our experimental protocol.

\subsubsection{Cross validation}

We applied a 5-fold cross validation.
We split the seed entities of each dataset into 5 sets $\mathcal{S}_i,~i\in\{1,2,3,4,5\}$, where each set contains the same number of seed entities. 
Each set $\mathcal{S}_i$ is successively used for testing, while $\mathcal{S}_{(i-1)}$ is used for validation, and the remaining sets are used for training.
To prevent over-fitting, we implement an early-stopping method based on the validation loss. We set the patience to 5 for models trained with 50 epochs, and to 20 for models trained with 200 epochs.

Such a splitting on seed entities at testing, guarantees that  we evaluate the ability of the model to generalize on unseen seed entities.
Additionally, some entities reached from these unseen seed entities are not seen during training either, as highlighted in Table~\ref{tab:rate-seen-qids}.
Such an experimental setup thus assesses the model's capability  to learn a relative similarity or dissimilarity between seed entities and reached entities, and to extrapolate it on unseen seed and reached entities. This extrapolation roots our zero-shot approach.

It can be noticed in Table~\ref{tab:rate-seen-qids} that test seed entities in Dataset 1 lead to more entities that were seen in training than in Dataset 2 (51-61\% instead of 10-21\%).
This is a direct consequence of the homogeneity of seed entities in Dataset 1.
Since all seed entities are from the Computer Science / Information Technology domain, the entities traversed during the expansion from each seed entity may overlap.
On the contrary, Dataset 2 involves heterogeneous seed entities from different domains, leading to different entities being traversed when expanding from each seed entity.

\begin{table}
    \begin{tabular}{l|rrrrr}
         \toprule
         \multicolumn{1}{c}{Dataset} & \multicolumn{1}{|c}{Fold 1} & \multicolumn{1}{c}{Fold 2} & \multicolumn{1}{c}{Fold 3} & \multicolumn{1}{c}{Fold 4} & \multicolumn{1}{c}{Fold 5} \\
         \midrule
         Dataset 1 & 56.51 & 59.61 & 51.72 & 55.76 & 61.19 \\
         Dataset 2 & 20.50 & 21.49 & 12.36 & 17.04 & 10.05 \\
         \bottomrule
    \end{tabular}
    \caption{Percentage of entities reached when testing and already seen when training, for each fold and dataset.}
    \label{tab:rate-seen-qids}
\end{table}

\subsubsection{Transfer learning}

We also tested our model in a transfer learning setting. 
We trained our model on labeled decisions of Dataset 1 and tested it by traversing the neighborhood of seed entities in Dataset 2.

\subsection{Models}
\label{subsection:models}

We compare our proposed analogy-based model to the following baseline models: MLP, LSTM, Random Forest, SVM, depth pruning, and threshold pruning~\cite{jarnacM22}.
We call \emph{analogy} the model that does not considers paths (Equation~\eqref{eq:analogy}), and \emph{path analogy} the model that consider paths (Equation~\eqref{eq:path-analogy}).

Note that the dimension of the pre-trained embeddings of Wikidata is 200.
Some parameters are used by several models and are detailed below:

\begin{description}

\item[Batch size] We set the batch size do 32.

\item[Optimizer] We use the Adam optimizer.
 
\item[Embedding] We consider the embeddings $\mathcal{E}_1$ and $\mathcal{E}_2$, as proposed in \cite{jarnacM22} and explained in Subsection~\ref{subsection:datasets}.

\item[Concatenation] Consider a pair $(e_s, e_r)$ formed by a seed entity and an entity reached. To decide whether to keep or prune $e_r$, some models can take as input the horizontal concatenation of the embeddings of $e_s$ and $e_r$ (called \textit{horizontal}) or their difference (called \textit{translation}).

\item[Zero padding] We consider three zero-padding methods: \textit{before}, \textit{between}, and \textit{after}, as detailed in Subsubsection~\ref{subsubsection:model}.

\item[Learning rate] We test with learning rates $\in \{0.0001$, $0.001$, $0.01\}$.

\item[Dropout rate] We test with dropout rates $\in \{0, 0.3, 0.5\}$.

\item[Path length] We consider paths of length $\in \{3, 4, 5\}$.

\item[Analogy configuration] We consider the three configurations for valid and invalid analogies $\mathcal{C}_1$, $\mathcal{C}_2$, $\mathcal{C}_3$ presented in Subsubsection~\ref{subsubsection:formalization}.

\item[Number of filters] We test with $(n_1, n_2) \in \{(2, 1), (4, 2), (8, 4)$, $(16, 8), (32, 16), (64, 32), (128, 64), (256, 128)\}$.
\end{description}

The parameters used by the different considered models are given below, where
 we also describe specific parameters that are only applicable to one model.
\begin{description}
    \item[Analogy (A)] Batch size, optimizer, embedding, learning rate, drop\-out rate, analogy configuration, number of filters, $M \in \{5$, $10$, $15$, $20$, $50$, $100\}$, $N = 20$.
    \item[Path analogy (PA)] Batch size, optimizer, embedding, learning rate, dropout rate, zero padding, path length, analogy configuration, number of filters, $M \in \{5, 10, 15, 20, 50, 100\}$, and $N = 20$. 
    \item[SVM] Embedding, concatenation, and unlimited number of iterations.
    \item[Random Forest (RF)] Embedding, concatenation, and number of estimators $\in \{10, 50, 100, 150, 200, 250, 300, 400, 500 \}$.
    \item[MLP] Batch size, optimizer, embedding, concatenation, learning rate, dropout rate, and hidden layers $\in$ $\{(100)$, $(100, 50)$, $(100, 50, 25)$, $(200)$, $(200, 100)$, $(200, 100, 50)$, $(200, 100, 50, 25)\}$.
    \item[LSTM] Batch size, optimizer, embedding, learning rate, zero padding, path length, and number of units $\in \{50, 100, 150\}$.
    \item[Depth pruning (D)] Depth threshold $\in \llbracket1, 20\rrbracket$.
    \item[Threshold pruning (T)] Embedding, $\alpha \in \{1.0, 1.1, \dots, 2.0\}$, $\gamma = 20$, $\beta \in \{1.0, 1.1 \dots, 2.0\}$, and $\text{absolute degree} = 200$.
\end{description}
Note that for models with a non-zero dropout rate, we use Monte Carlo Dropout.

For all models except analogy-based models, we explored all combinations of different parameter values.
Given the important parameter space, for analogy-based models  we first fixed the embedding to $\mathcal{E}_1$, and the configuration to $\mathcal{C}_1$ on Dataset 2 in order to find the three best numbers of filters, the two best path lengths, the best zero padding method, and the best dropout rate.
We then experimented with this reduced parameter space on Dataset 1 and Dataset 2.

\subsection{Evaluation Metrics}
\label{subsection:metrics}

\begin{figure}
    \centering
    \includegraphics[scale=0.5]{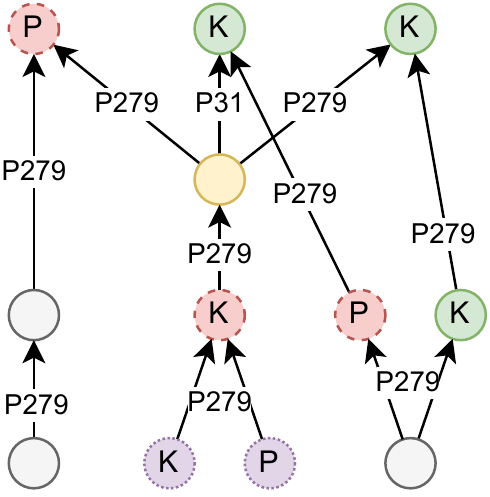}
    \caption[]{Example of an evaluation setting for a pruning model. Gold decisions are represented by the letters K (for keep) and P (for prune).
    \raisebox{-2pt}{\begin{tikzpicture}[scale=0.2]
        \node[draw=seedborder, circle, fill=seed]{} ;
    \end{tikzpicture}} represents the seed entity;
    \raisebox{-2pt}{\begin{tikzpicture}[scale=0.5]
        \node[draw=greenborder, circle, fill=green]{} ;
    \end{tikzpicture}} are entities kept by the model;
    \raisebox{-2pt}{\begin{tikzpicture}[scale=0.5]
        \node[draw=redborder, circle, dashed, fill=red]{} ;
    \end{tikzpicture}} are entities pruned by the model;
    \raisebox{-2pt}{\begin{tikzpicture}[scale=0.5]
        \node[draw=purpleborder, circle, dotted, fill=purple]{} ;
    \end{tikzpicture}} are entities unexplored by the model due to pruning decisions but that have a gold decision;
    \raisebox{-2pt}{\begin{tikzpicture}[scale=0.5]
        \node[draw=greyborder, circle, fill=grey]{} ;
    \end{tikzpicture}} represent unlabeled entities that are not considered in the evaluation.}
    \label{fig:pruning-evaluation}
\end{figure}

Figure~\ref{fig:pruning-evaluation} illustrates the expansion along the ontology hierarchy from a seed entity (\raisebox{-2pt}{\begin{tikzpicture}\node[draw=seedborder, circle, fill=seed]{} ;\end{tikzpicture}}) with a pruning model.
In such a setting, we only evaluate the model on nodes that are associated with a gold decision, \textit{i.e.}, nodes depicted by \raisebox{-2pt}{\begin{tikzpicture}[scale=0.5]\node[draw=greyborder, circle, fill=grey]{} ;\end{tikzpicture}} are not considered.
In this view, it should be noted that it is possible for the model to leave nodes with gold decisions unexplored due to erroneous pruning decisions higher in the hierarchy (\textit{i.e.}, nodes depicted by \raisebox{-2pt}{\begin{tikzpicture}[scale=0.5]
        \node[draw=purpleborder, circle, dotted, fill=purple]{} ;
    \end{tikzpicture}}).
To take into account these various cases in our evaluation, we use the following metrics:

\begin{equation*}
    \text{Precision} = \cfrac{\raisebox{-6pt}{\begin{tikzpicture}[scale=0.5]\node[draw=greenborder, circle, fill=green,font=\scriptsize]{K} ;\end{tikzpicture}}}{\raisebox{-6pt}{\begin{tikzpicture}[scale=0.5]\node[draw=greenborder, circle, fill=green,font=\scriptsize]{K} ;\end{tikzpicture}} ~+~ \raisebox{-6pt}{\begin{tikzpicture}[scale=0.5]\node[draw=greenborder, circle, fill=green,font=\scriptsize]{P} ;\end{tikzpicture}}} \hspace{1em} 
    \text{Recall} = \cfrac{\raisebox{-6pt}{\begin{tikzpicture}[scale=0.5]\node[draw=greenborder, circle, fill=green,font=\scriptsize]{K} ;\end{tikzpicture}}}{\raisebox{-6pt}{\begin{tikzpicture}[scale=0.5]\node[draw=greenborder, circle, fill=green,font=\scriptsize]{K} ;\end{tikzpicture}} ~+~ \raisebox{-6pt}{\begin{tikzpicture}[scale=0.5]\node[draw=purpleborder, circle, dotted, fill=purple,font=\scriptsize]{K} ;\end{tikzpicture}} ~+~ \raisebox{-6pt}{\begin{tikzpicture}[scale=0.5]\node[draw=redborder, circle, dashed, fill=red,font=\scriptsize]{K} ;\end{tikzpicture}}}
\end{equation*}
\begin{equation*}
    \text{Accuracy} = \cfrac{\raisebox{-6pt}{\begin{tikzpicture}[scale=0.5]\node[draw=greenborder, circle, fill=green, font=\scriptsize]{K} ;\end{tikzpicture}} ~+~ \raisebox{-6pt}{\begin{tikzpicture}[scale=0.5]\node[draw=redborder, circle, dashed, fill=red, font=\scriptsize]{P} ;\end{tikzpicture}}}{\raisebox{-6pt}{\begin{tikzpicture}[scale=0.5]\node[draw=greenborder, circle, fill=green, font=\scriptsize]{K} ;\end{tikzpicture}} ~+~ \raisebox{-6pt}{\begin{tikzpicture}[scale=0.5]\node[draw=greenborder, circle, fill=green, font=\scriptsize]{P} ;\end{tikzpicture}} ~+~ \raisebox{-6pt}{\begin{tikzpicture}[scale=0.5]\node[draw=redborder, circle, dashed, fill=red, font=\scriptsize]{K} ;\end{tikzpicture}} ~+~ \raisebox{-6pt}{\begin{tikzpicture}[scale=0.5]\node[draw=redborder, circle, dashed, fill=red, font=\scriptsize]{P} ;\end{tikzpicture}} ~+~ \raisebox{-6pt}{\begin{tikzpicture}[scale=0.5]\node[draw=purpleborder, circle, dotted, fill=purple, font=\scriptsize]{K} ;\end{tikzpicture}} ~+~ \raisebox{-6pt}{\begin{tikzpicture}[scale=0.5]\node[draw=purpleborder, circle, dotted, fill=purple, font=\scriptsize]{P} ;\end{tikzpicture}}}
\end{equation*}
\begin{equation*}
\text{F1} = 2 \times \cfrac{\text{Precision} \times \text{Recall}}{\text{Precision} + \text{Recall}}    
\end{equation*}

This corresponds to a binary classification in which the keeping decision is the positive class and the pruning decision is the negative class.

\subsection{Results}

We introduce our results in this section following the two setups described in Section~\ref{section:experiments}.
We further discuss them in Section~\ref{section:discussion}.

\subsubsection{Cross validation} 
\begin{figure}
     \centering
     \begin{subfigure}[b]{\linewidth}
         \centering
         \includegraphics[width=0.6\textwidth]{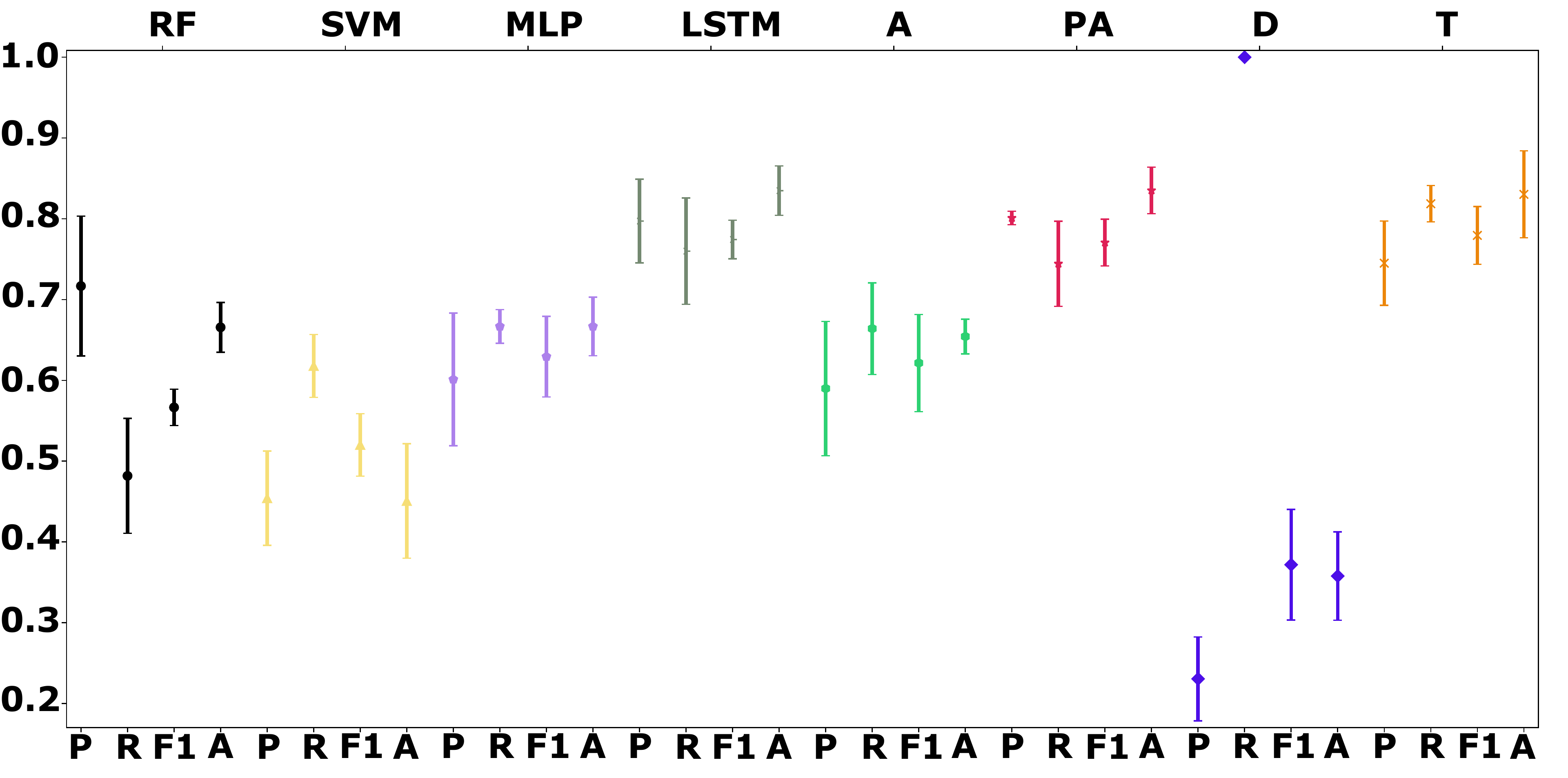}
         \caption{Dataset 1}
         \label{subfigure:metrics-dataset1}
     \end{subfigure}
     
     \begin{subfigure}[b]{\linewidth}
         \centering
         \includegraphics[width=0.6\textwidth]{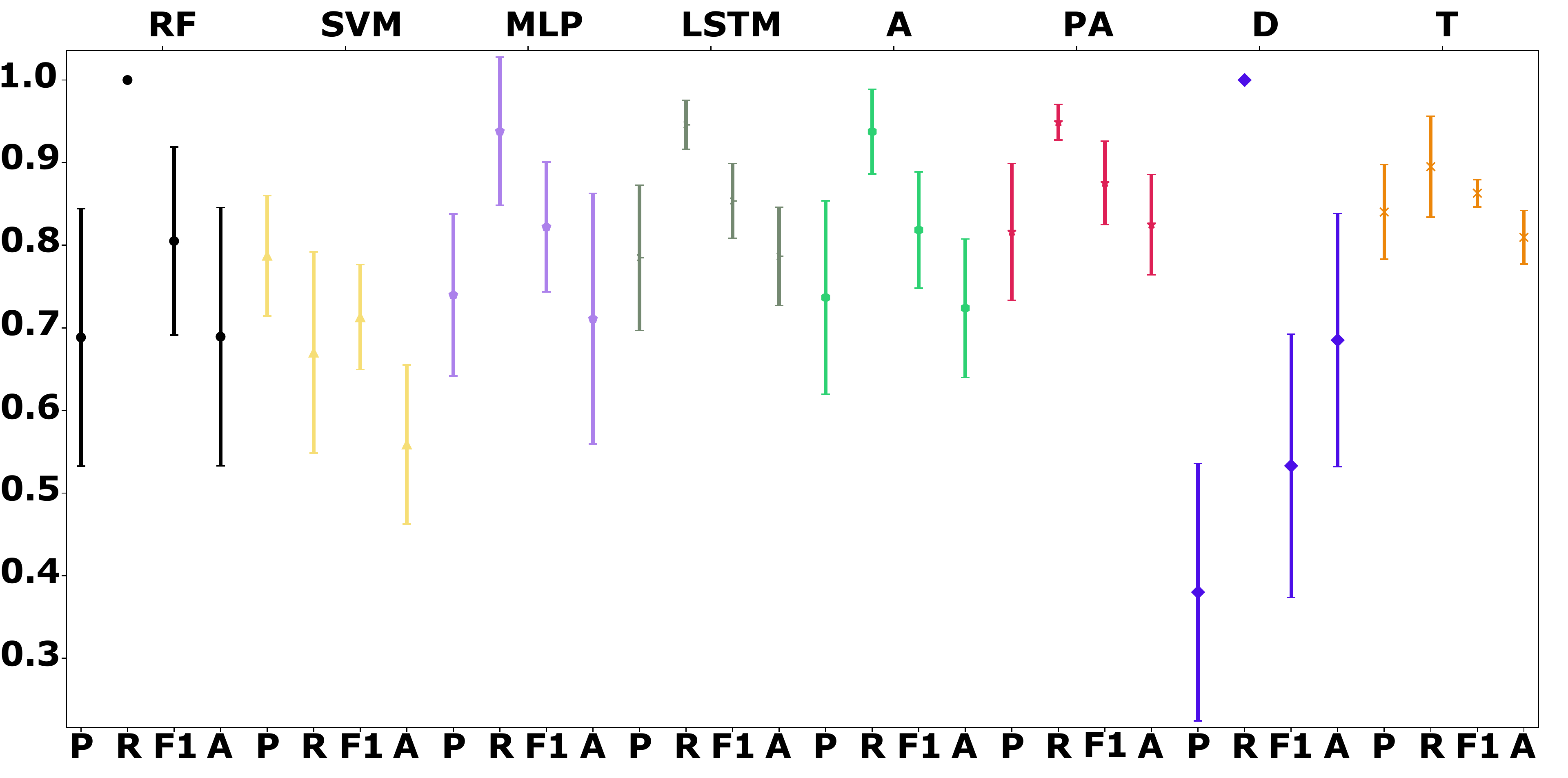}
         \caption{Dataset 2}
         \label{subfigure:metrics-dataset2}
     \end{subfigure}
    \caption{Error plot depicting the precision (P), recall (R), F1-score (F1), and accuracy (A) of each model on Dataset 1 and Dataset 2.}
    \label{fig:pruning-performance}
\end{figure}

We present in Table~\ref{tab:pruning-results} the performance of the different models on the task of keeping relevant entities and pruning irrelevant ones on Dataset 1 and Dataset 2.
Figure~\ref{fig:pruning-performance} depicts these results with error plots to better assess the variability or stability of each model.
Note that we present the results of the reference Threshold whose decisions were labeled to build the datasets.
However, we do not use them to draw comparisons and conclusions because of the bias that would constitute using such results in both the dataset building and evaluation processes.

For each model, Table~\ref{tab:pruning-results} and Figure~\ref{fig:pruning-performance} only present the best results in terms of F1-score (primary criterion) and accuracy (secondary criterion) that were obtained when exploring the parameter space.
The best parameters were the following:
\begin{description}
    \item[Analogy] Embedding $\mathcal{E}_1$, $(n_1, n_2) = (16, 8)$, $\text{dropout} = 0.5$, configuration $\mathcal{C}_2$ and $\text{learning rate} = 0.001$ (for Dataset 1), configuration $\mathcal{C}_1$ and $\text{learning rate} = 0.01$ (for Dataset 2).
    \item[Path analogy] Embedding $\mathcal{E}_1$, configuration $\mathcal{C}_1$, $\text{learning rate} = 0.001$, $\text{zero-padding} = \text{between}$, $\text{path length} = 4$, $(n_1, n_2) = (16, 8)$, and $\text{dropout} = 0$ (for Dataset 1), $\text{path length} = 3$, $(n_1, n_2) = (4, 2)$, and $\text{dropout} = 0.3$ (for Dataset 2).
    \item[SVM] Embedding $\mathcal{E}_1$, $\text{concatenation} = \text{horizontal}$ (for Dataset 1), and $\text{concatenation} = \text{translation}$ (for Dataset 2).
    \item[Random Forest] $\text{concatenation} = \text{horizontal}$, embedding $\mathcal{E}_2$ and 200 estimators (for Dataset 1), embedding $\mathcal{E}_1$ and 300 estimators (for Dataset 2).
    \item[MLP] $\text{hidden layers} = (200, 100, 50)$, embedding $\mathcal{E}_1$, concatenation $= \text{horizontal}$, $\text{learning rate} = 0.001$, $\text{dropout} = 0$ (for Dataset 1), and Embedding $\mathcal{E}_2$, $\text{concatenation} = \text{translation}$, $\text{learning rate} = 0.01$, $\text{dropout} = 0.3$ (for Dataset 2).
    \item[LSTM] Embedding $\mathcal{E}_1$, $\text{zero-padding} = \text{before}$, number of units $= 150$, $\text{path length} = 5$ and $\text{learning rate} = 0.01$ (for Dataset 1), $\text{path length} = 3$ and $\text{learning rate} = 0.001$ (for Dataset 2).
    \item[Depth pruning] Depth threshold $= 4$ (for Dataset 1), and depth threshold $ = 3$ (for Dataset 2).
    \item[Threshold pruning] $\alpha = 1.0$, embedding $\mathcal{E}_1$, $\beta = 2.0$ (for Dataset 1), $\beta = 1.8$ (for Dataset 2).
\end{description}

\begin{table*}
    \begin{adjustbox}{width={\textwidth},totalheight={0.2\textheight},keepaspectratio}%
    \begin{tabular}{l|rrrr|rrrr}
         \toprule
         \multicolumn{1}{c}{\multirow{2}{*}{Model}} & \multicolumn{4}{|c}{Dataset 1} & \multicolumn{4}{|c}{Dataset 2} \\
         \cmidrule{2-9}
         \multicolumn{1}{c}{} & \multicolumn{1}{|c}{P} & \multicolumn{1}{c}{R} & \multicolumn{1}{c}{F1} & \multicolumn{1}{c}{ACC} & \multicolumn{1}{|c}{P} & \multicolumn{1}{c}{R} & \multicolumn{1}{c}{F1} & \multicolumn{1}{c}{ACC} \\
         \midrule
         Random Forest & $71.68 \pm 8.66$ & $48.18 \pm 7.10$ & $56.66 \pm 2.25$ & $66.57 \pm 3.09$ & $68.85 \pm 15.59$ & $\textbf{100.00} \pm \textbf{0.00}$ & $80.50 \pm 11.39$ & $68.93 \pm 15.62$\\
         SVM & $45.43 \pm 5.84$ & $61.79 \pm 3.90$ & $52.02 \pm 3.86$ & $45.08 \pm 7.09$ & $78.73 \pm 7.29$ & $67.01 \pm 12.18$ & $71.29 \pm 6.34$ & $55.88 \pm 9.63$\\
         MLP & $60.12 \pm 8.20$ & $66.68 \pm 2.09$ & $62.94 \pm 4.98$ & $66.68 \pm 3.62$ & $73.99 \pm 9.80$ & $93.80 \pm 8.96$ & $82.22 \pm 7.86$ & $71.10 \pm 15.18$\\
         LSTM & $\underline{79.72 \pm 5.17}$ & $\textbf{76.00} \pm \textbf{6.59}$ & $\textbf{77.43} \pm \textbf{2.38}$ & $\underline{83.48 \pm 3.05}$ & $\underline{78.49 \pm 8.80}$ & $94.58 \pm 2.96$ & $\underline{85.36 \pm 4.53}$ & $\underline{78.66 \pm 5.95}$\\
         Analogy & $58.99 \pm 8.31$ & $66.41 \pm 5.67$ & $62.15 \pm 6.01$ & $65.43 \pm 2.16$ & $73.67 \pm 11.71$ & $93.75 \pm 5.12$ & $81.85 \pm 7.03$ & $72.39 \pm 8.38$\\
         Path analogy & $\textbf{80.10} \pm \textbf{0.84}$ & $\underline{74.44 \pm 5.28}$ & $\underline{77.06 \pm 2.89}$ & $\textbf{83.51} \pm \textbf{2.87}$ & $\textbf{81.63} \pm \textbf{8.27}$ & $\underline{94.90 \pm 2.16}$ & $\textbf{87.54} \pm \textbf{5.05}$ & $\textbf{82.50} \pm \textbf{6.07}$\\
         \midrule
         Depth & $ 23.06\pm 5.19$ & $100.00 \pm 0.00$ & $37.19 \pm 6.85$ & $35.78 \pm 5.48$ & $38.01 \pm 15.57$ & $100.00 \pm 0.00$ & $53.30 \pm 15.92$ & $68.51 \pm 15.30$\\
         \midrule
         \rowcolor{darkgrey}Threshold & $74.50 \pm 5.22$ & $81.86 \pm 2.25$ & $77.94 \pm 3.58$ & $83.02 \pm 5.38$ & $84.02 \pm 5.72$ & $89.51 \pm 6.12$ & $86.30 \pm 1.66$ & $80.96 \pm 3.24$\\
         \bottomrule
    \end{tabular}
    \end{adjustbox}
    \caption{Pruning evaluation results on Dataset 1 and Dataset 2, with the parameters leading to the best results for each model. P stands for average precision, R stands for average recall, F1 stands for average F1-score, and ACC stands for average accuracy. 
    The best results are in bold and we underline the second best result. Also, we decided to present the reference threshold results used in the construction of both datasets.}
    \label{tab:pruning-results}
\end{table*}

\subsubsection{Transfer learning}

For our transfer learning setting, we used the best parameters found during the cross-validation on Dataset 1.
We trained each model on 80\% of Dataset 1, using 20\% as validation for early-stopping.
We then tested the trained models on Dataset 2.
Results are presented in Table~\ref{tab:transfer-results}.

\begin{table}
    \begin{tabular}{l|rrrr}
         \toprule
         \multicolumn{1}{c}{Model} & \multicolumn{4}{|c}{Dataset 1 $\to$ 2}\\
         \cmidrule{2-5}
         \multicolumn{1}{c}{} & \multicolumn{1}{|c}{P} & \multicolumn{1}{c}{R} & \multicolumn{1}{c}{F1} & \multicolumn{1}{c}{ACC} \\
         \midrule
         Random Forest & $83.09$ & $20.40$ & $32.75$ & $33.06$ \\
         SVM & $64.62$ & $ 57.04$ & $60.59$ & $43.97$ \\
         MLP & $69.90$ & $51.99$ & $59.63$ & $48.26$ \\
         LSTM & $\textbf{92.83}$ & $\underline{74.73}$ & $\underline{82.80}$ & $\underline{80.04}$ \\
         Analogy & $74.95$ & $64.26$ & $69.19$ & $59.86$ \\
         Path analogy & $\underline{91.49}$ & $\textbf{83.39}$ & $\textbf{87.25}$ & $\textbf{84.33}$ \\
         \bottomrule
    \end{tabular}
    \caption{Transfer results obtained by training on Dataset 1 and testing on Dataset 2. P stands for average precision, R stands for average recall, F1 stands for average F1-score, and ACC stands for average accuracy. The best results are in bold and the second best  are underlined.}
    \label{tab:transfer-results}
\end{table}

\section{Discussion}
\label{section:discussion}

Table~\ref{tab:pruning-results}  highlights that depth in the ontology hierarchy cannot be used to keep relevant entities and prune irrelevant ones. 
Here, with a depth of 3-4, we obtain a perfect recall but a low precision.
Using greater thresholds does not change results while lower ones would reduce the recall in favor of the precision. 
This was expected, especially in a collaborative and generic knowledge graph such as Wikidata, since different communities may have different granular representations of knowledge.
To tackle this issue, one would need to specify different depth thresholds depending on the seed entity domains.
Additionally, not all subclasses of an interesting class may be of interest w.r.t. a seed entity, especially in case of errors in the ontology hierarchy (\textit{e.g.}, erroneous subclass edges, misinterpretation of the subclass semantics by users, introduction of cycles). 
Such results motivate the use of classifiers to learn a relative similarity and dissimilarity between reached entities and seed entities, based on labeled examples given by a user (\textit{e.g.}, an expert, the KG owner).

Regarding classifier performance, the LSTM and path analogy models are the best performing models, which outlines the importance of paths leading to entities to decide on their relevance.
Figure~\ref{fig:pruning-performance} indicates that the LSTM and path analogy models are the more stable, with the path analogy model being particularly stable on Dataset 1 for the precision metric.
On this dataset, the path analogy model obtains the best results in precision and accuracy whereas the LSTM has the best F1-score and recall.
However, it should be noted that scores are close.
On Dataset 2, the path analogy model outperforms the LSTM by about 3 points in precision, 2 points in F1-score, and 4 points in accuracy, while obtaining a similar recall.
Recall that Dataset 1 is based on a set of homogeneous seed entities from the Computer Science / Information Technology domain whereas Dataset 2 mixes heterogenous domains such as food, sport, and science.
Dataset 1 may thus provide less diversity for models to correctly learn similarity and dissimilarity between reached entities and seed entities. 
Additionally, such an homogeneity may also entail a fuzzy keeping/pruning boundary.
To illustrate, starting from a network protocol of a specific layer of the OSI model, a protocol from another OSI layer was manually labeled with a pruning decision.
Such a very fine-grained decision may be difficult to capture by models.
To better reflect the performance of each model, we also provide their number of trainable parameters in Table~\ref{tab:trainable-parameters}.
It is then striking that the path analogy model obtains close or better performance than the LSTM with 150 to 800 times fewer parameters.
This global evaluation, taking into account the performance measured with traditional metrics as well as the number of trainable parameters, shows the superiority of our proposed analogy-based model.

\begin{table}
    \begin{tabular}{l|rr}
         \toprule
         \multicolumn{1}{c}{Model} & \multicolumn{1}{|c}{Dataset 1} & \multicolumn{1}{c}{Dataset 2} \\
         \midrule
         LSTM & 210,751 & 210,751\\
         MLP & 105,401 & 65,401\\
         Analogy  & 1,369 & 1,369\\
         Path analogy & 1,401 & 251\\
         \bottomrule
    \end{tabular}
    \caption{Number of trainable parameters for each model.}
    \label{tab:trainable-parameters}
\end{table}

\begin{table*}[h!]
    \begin{adjustbox}{width={\textwidth},totalheight={0.2\textheight},keepaspectratio}%
    \begin{tabular}{ll|rrrr|rrrr}
        \toprule
         \multicolumn{2}{c}{} & \multicolumn{4}{|c}{Dataset 1} & \multicolumn{4}{|c}{Dataset 2} \\
         \cmidrule{3-10}
         \multicolumn{2}{c}{} & \multicolumn{1}{|c}{P} & \multicolumn{1}{c}{R} & \multicolumn{1}{c}{F1} & \multicolumn{1}{c}{ACC} & \multicolumn{1}{|c}{P} & \multicolumn{1}{c}{R} & \multicolumn{1}{c}{F1} & \multicolumn{1}{c}{ACC} \\
         \midrule
         \multirow{2}{*}{LSTM} & Unseen entities & $\textbf{73.90} \pm \textbf{5.97}$ & $\textbf{69.69} \pm \textbf{7.90}$ & $71.33 \pm 4.73$ & $78.06 \pm 6.15$ & $75.44 \pm 8.99$ & $94.21 \pm 2.99$ & $83.33 \pm 4.72$ & $76.35 \pm 6.37$ \\ 
         & Seen entities & $84.63 \pm 6.80$ & $\textbf{80.95} \pm \textbf{5.51}$ & $\textbf{82.32} \pm \textbf{1.29}$ & $\textbf{87.55} \pm \textbf{1.83}$ & $94.62 \pm 1.71$ & $\textbf{95.51} \pm \textbf{3.99}$ & $95.02 \pm 2.14$ & $91.71 \pm 3.39$ \\
         \midrule
         \multirow{2}{*}{Path analogy} & Unseen entities & $73.82 \pm 3.62$ & $69.44 \pm 6.09$ & $\textbf{71.36} \pm \textbf{3.64}$ & $\textbf{78.20} \pm \textbf{4.78}$ & $\textbf{78.84} \pm \textbf{8.34}$ & $\textbf{94.46} \pm \textbf{2.83}$ & $\textbf{85.67} \pm \textbf{5.01}$ & $\textbf{80.49} \pm \textbf{5.97}$ \\ 
         & Seen entities & $\textbf{85.20} \pm \textbf{2.10}$ & $78.25 \pm 5.20$ & $81.47 \pm 3.01$ & $87.41 \pm 2.05$ & $\textbf{96.07} \pm \textbf{2.47}$ & $94.90 \pm 5.45$ & $\textbf{95.44} \pm \textbf{3.82}$ & $\textbf{93.15} \pm \textbf{4.98}$ \\
         \bottomrule
    \end{tabular}
    \end{adjustbox}
    \caption{Performance of the LSTM and path analogy models depending on whether entities reached when testing where seen or unseen in training. P stands for average precision, R stands for average recall, F1 stands for average F1-score, and ACC stands for average accuracy.
    Best results for each category (\textit{i.e.}, seen or unseen) are in bold.}
    \label{tab:seen-unseen}
\end{table*}

Table~\ref{tab:transfer-results} shows the performance of the compared models on the transfer learning setting. 
Again, it appears that the LSTM and path-analogy models are the two best performing models. 
While the LSTM obtains a better precision, the path-analogy model outperforms on recall, F1-score, and accuracy by 4 to 9 points. 
To better assess the generalization capability of these two models, we provide in Table~\ref{tab:seen-unseen} the breakdown of results from Table~\ref{tab:pruning-results} depending on whether the reached entities when testing were seen or unseen during training.
As can be expected, both models perform less on unseen entities.
We notice that their scores are similar on Dataset~1 whereas the path analogy model outperforms the LSTM on both unseen and seen entities in Dataset~2. 
Recall that Dataset~2 contains much more unseen entities in testing that Dataset~1 (Table~\ref{tab:rate-seen-qids}).
These results thus demonstrate the higher generalization capability of our proposed analogy-based model. 
We posit that the formalization of analogical quadruples and the use of a CNN lead the model to learn to compute and compare \emph{relative} similarities and dissimilarities between the two pairs in a quadruple.
In turn, this leads to better extrapolation capabilities to decide on an unseen pair when compared to a seen pair within an analogical quadruple.
Consequently, we think analogy-based model are well-suited for such zero-shot settings.

To extend our approach, we could envision to test our model in a few-shot setting by having some labeled decisions to train on for seed entities considered in testing. 
In a real-world use-case scenario, this would correspond to asking experts to label a few neighbors of each seed entity before performing the expansion along the ontology hierarchy.
This could be of interest to further test the extrapolation capability of our analogy-based model.
However, we believe that in a real-world scenario, experts would rather label as many neighbors as possible of some seed entities and expect the model to extrapolate on new seed entities, hence our focus on the zero-shot setting.
As aforementioned, some pruning decisions are motivated by errors in the ontology hierarchy of Wikidata, which is known to contain to be potentially noisy~\cite{shenoy22}\footnote{See also \url{https://commons.wikimedia.org/wiki/File:WikidataCon_2021_-_Overview_of_ontology_issues.pdf}.}.
Another extension of our approach thus consists in applying it to ontology maintenance.

Regarding our model, we leverage KG embeddings pre-trained with a translational model.
However, there exist several types of KG embedding models, such as translational, complex, Gaussian or Graph Neural Network-based ones~\cite{ji2022}.
It would thus be interesting to evaluate which types of embedding models are better suited to serve an analogy-based model.
Additionally, instead of using frozen KG embeddings previously learned on a specific task, we could envision learning simultaneously the graph embeddings and the CNN layers, similarly to what was done in~\cite{marquer22} with character and word embeddings.
Finally, it is noteworthy that our work does not rely on Large Language Models (LLMs) such as BERT~\cite{devlinCLT19}. 
This purposely allows us to assess if the structure of the graph provides enough information to learn useful embeddings for selecting relevant entities. 
Future research directions could involve enriching our approach with LLMs, while raising additional issues to face such as noisy labels or homonyms.
To illustrate, the entity ``role'' can be a part played by a performer\footnote{\url{https://www.wikidata.org/wiki/Q1707847}} or an identity of an item in relation to another specified item\footnote{\url{https://www.wikidata.org/wiki/Q4897819}}.

\section{Conclusion}
\label{section:conclusion}

In this paper, we considered the task of bootstrapping a knowledge graph (KG) by selecting relevant entities in the neighborhood of seed entities of interest in a generic KG.
We proposed an analogy-based model to keep or prune neighors of seed entities in a zero-shot setting and two labeled datasets to evaluate models on this task.
Compared with standard classifiers, our model outperformed while presenting a drastically lower number of parameters. 
Additionally, it showed better extrapolation capabilities in zero-shot and transfer learning settings. 
Such results advocate for the further study of analogy-based models in tasks related to the KG lifecycle or requiring extrapolation capabilities, which we will address  in future work.

\begin{acks}
This work is supported by the AT2TA project (\url{https://at2ta.loria.fr/}) funded by the French National Research Agency (``Agence Nationale de la Recherche'' -- ANR) under grant ANR-22-CE23-0023. 
\end{acks}

\bibliographystyle{ACM-Reference-Format}
\balance
\bibliography{bibliography}


\begin{thebibliography}{51}


\ifx \showCODEN    \undefined \def \showCODEN     #1{\unskip}     \fi
\ifx \showDOI      \undefined \def \showDOI       #1{#1}\fi
\ifx \showISBNx    \undefined \def \showISBNx     #1{\unskip}     \fi
\ifx \showISBNxiii \undefined \def \showISBNxiii  #1{\unskip}     \fi
\ifx \showISSN     \undefined \def \showISSN      #1{\unskip}     \fi
\ifx \showLCCN     \undefined \def \showLCCN      #1{\unskip}     \fi
\ifx \shownote     \undefined \def \shownote      #1{#1}          \fi
\ifx \showarticletitle \undefined \def \showarticletitle #1{#1}   \fi
\ifx \showURL      \undefined \def \showURL       {\relax}        \fi
\providecommand\bibfield[2]{#2}
\providecommand\bibinfo[2]{#2}
\providecommand\natexlab[1]{#1}
\providecommand\showeprint[2][]{arXiv:#2}

\bibitem[Afantenos et~al\mbox{.}(2021)]%
        {afantenos.2021}
\bibfield{author}{\bibinfo{person}{Stergos~D. Afantenos},
  \bibinfo{person}{Tarek Kunze}, \bibinfo{person}{Suryani Lim},
  \bibinfo{person}{Henri Prade}, {and} \bibinfo{person}{Gilles Richard}.}
  \bibinfo{year}{2021}\natexlab{}.
\newblock \showarticletitle{Analogies Between Sentences: Theoretical Aspects -
  Preliminary Experiments}. In \bibinfo{booktitle}{\emph{Symbolic and
  Quantitative Approaches to Reasoning with Uncertainty - 16th European
  Conference, {ECSQARU} 2021, Prague, Czech Republic, September 21-24, 2021,
  Proceedings}} \emph{(\bibinfo{series}{Lecture Notes in Computer Science},
  Vol.~\bibinfo{volume}{12897})}. \bibinfo{publisher}{Springer},
  \bibinfo{pages}{3--18}.
\newblock
\urldef\tempurl%
\url{https://doi.org/10.1007/978-3-030-86772-0_1}
\showDOI{\tempurl}


\bibitem[Afantenos et~al\mbox{.}(2022)]%
        {afantenos.2022}
\bibfield{author}{\bibinfo{person}{Stergos~D. Afantenos},
  \bibinfo{person}{Suryani Lim}, \bibinfo{person}{Henri Prade}, {and}
  \bibinfo{person}{Gilles Richard}.} \bibinfo{year}{2022}\natexlab{}.
\newblock \showarticletitle{Theoretical Study and Empirical Investigation of
  Sentence Analogies}. In \bibinfo{booktitle}{\emph{Proceedings of the Workshop
  on the Interactions between Analogical Reasoning and Machine Learning
  (International Joint Conference on Artificial Intelligence - European
  Conference on Artificial Intelligence {(IJAI-ECAI} 2022)), Vienna, Austria,
  July 23, 2022}} \emph{(\bibinfo{series}{{CEUR} Workshop Proceedings},
  Vol.~\bibinfo{volume}{3174})}. \bibinfo{publisher}{CEUR-WS.org},
  \bibinfo{pages}{15--28}.
\newblock
\urldef\tempurl%
\url{https://ceur-ws.org/Vol-3174/paper2.pdf}
\showURL{%
\tempurl}


\bibitem[Alsaidi et~al\mbox{.}(2021)]%
        {detecting:2021:alsaidi}
\bibfield{author}{\bibinfo{person}{Safa Alsaidi}, \bibinfo{person}{Amandine
  Decker}, \bibinfo{person}{Puthineath Lay}, \bibinfo{person}{Esteban Marquer},
  \bibinfo{person}{Pierre{-}Alexandre Murena}, {and} \bibinfo{person}{Miguel
  Couceiro}.} \bibinfo{year}{2021}\natexlab{}.
\newblock \showarticletitle{A Neural Approach for Detecting Morphological
  Analogies}. In \bibinfo{booktitle}{\emph{8th {IEEE} International Conference
  on Data Science and Advanced Analytics, {DSAA} 2021, Porto, Portugal, October
  6-9, 2021}}. \bibinfo{publisher}{{IEEE}}, \bibinfo{pages}{1--10}.
\newblock
\urldef\tempurl%
\url{https://doi.org/10.1109/DSAA53316.2021.9564186}
\showDOI{\tempurl}


\bibitem[Babayeva et~al\mbox{.}(2022)]%
        {babayeva2022}
\bibfield{author}{\bibinfo{person}{Gulkhara Babayeva}, \bibinfo{person}{Kaie
  Maennel}, {and} \bibinfo{person}{Olaf~Manuel Maennel}.}
  \bibinfo{year}{2022}\natexlab{}.
\newblock \showarticletitle{Building an Ontology for Cyber Defence Exercises}.
  In \bibinfo{booktitle}{\emph{{IEEE} European Symposium on Security and
  Privacy, EuroS{\&}P 2022 - Workshops, Genoa, Italy, June 6-10, 2022}}.
  \bibinfo{publisher}{{IEEE}}, \bibinfo{pages}{423--432}.
\newblock
\urldef\tempurl%
\url{https://doi.org/10.1109/EuroSPW55150.2022.00050}
\showDOI{\tempurl}


\bibitem[Cai et~al\mbox{.}(2018)]%
        {caiZC18}
\bibfield{author}{\bibinfo{person}{Hongyun Cai}, \bibinfo{person}{Vincent~W.
  Zheng}, {and} \bibinfo{person}{Kevin~Chen{-}Chuan Chang}.}
  \bibinfo{year}{2018}\natexlab{}.
\newblock \showarticletitle{A Comprehensive Survey of Graph Embedding:
  Problems, Techniques, and Applications}.
\newblock \bibinfo{journal}{\emph{{IEEE} Transactions on Knowledge and Data
  Engineering}} \bibinfo{volume}{30}, \bibinfo{number}{9}
  (\bibinfo{year}{2018}), \bibinfo{pages}{1616--1637}.
\newblock
\urldef\tempurl%
\url{https://doi.org/10.1109/TKDE.2018.2807452}
\showDOI{\tempurl}


\bibitem[Chollet(2019)]%
        {Chollet}
\bibfield{author}{\bibinfo{person}{Fran{\c{c}}ois Chollet}.}
  \bibinfo{year}{2019}\natexlab{}.
\newblock \showarticletitle{On the Measure of Intelligence}.
\newblock \bibinfo{journal}{\emph{CoRR}}  \bibinfo{volume}{abs/1911.01547}
  (\bibinfo{year}{2019}).
\newblock


\bibitem[Couceiro et~al\mbox{.}(2017)]%
        {extend-boolean:2017:couceiro}
\bibfield{author}{\bibinfo{person}{Miguel Couceiro}, \bibinfo{person}{Nicolas
  Hug}, \bibinfo{person}{Henri Prade}, {and} \bibinfo{person}{Gilles Richard}.}
  \bibinfo{year}{2017}\natexlab{}.
\newblock \showarticletitle{Analogy-preserving functions: {A} way to extend
  Boolean samples}. In \bibinfo{booktitle}{\emph{Proceedings of the
  Twenty-Sixth International Joint Conference on Artificial Intelligence,
  {IJCAI} 2017, Melbourne, Australia, August 19-25, 2017}}.
  \bibinfo{publisher}{ijcai.org}, \bibinfo{pages}{1575--1581}.
\newblock
\urldef\tempurl%
\url{https://doi.org/10.24963/ijcai.2017/218}
\showDOI{\tempurl}


\bibitem[Couceiro et~al\mbox{.}(2018)]%
        {boolean-analogy:2018:couceiro}
\bibfield{author}{\bibinfo{person}{Miguel Couceiro}, \bibinfo{person}{Nicolas
  Hug}, \bibinfo{person}{Henri Prade}, {and} \bibinfo{person}{Gilles Richard}.}
  \bibinfo{year}{2018}\natexlab{}.
\newblock \showarticletitle{Behavior of Analogical Inference w.r.t. Boolean
  Functions}. In \bibinfo{booktitle}{\emph{Proceedings of the Twenty-Seventh
  International Joint Conference on Artificial Intelligence, {IJCAI} 2018, July
  13-19, 2018, Stockholm, Sweden}}. \bibinfo{publisher}{ijcai.org},
  \bibinfo{pages}{2057--2063}.
\newblock
\urldef\tempurl%
\url{https://doi.org/10.24963/ijcai.2018/284}
\showDOI{\tempurl}


\bibitem[Couceiro and Lehtonen(2023)]%
        {galois-analogical-clf:2023:couceiro-lehtonen}
\bibfield{author}{\bibinfo{person}{Miguel Couceiro} {and}
  \bibinfo{person}{Erkko Lehtonen}.} \bibinfo{year}{2023}\natexlab{}.
\newblock \showarticletitle{{Galois theory for analogical classifiers}}.
\newblock \bibinfo{journal}{\emph{AMAI}} (\bibinfo{year}{2023}).
\newblock
\urldef\tempurl%
\url{https://doi.org/10.1007/s10472-023-09833-6}
\showDOI{\tempurl}


\bibitem[Devlin et~al\mbox{.}(2019)]%
        {devlinCLT19}
\bibfield{author}{\bibinfo{person}{Jacob Devlin}, \bibinfo{person}{Ming{-}Wei
  Chang}, \bibinfo{person}{Kenton Lee}, {and} \bibinfo{person}{Kristina
  Toutanova}.} \bibinfo{year}{2019}\natexlab{}.
\newblock \showarticletitle{{BERT:} Pre-training of Deep Bidirectional
  Transformers for Language Understanding}. In
  \bibinfo{booktitle}{\emph{Proceedings of the 2019 Conference of the North
  American Chapter of the Association for Computational Linguistics: Human
  Language Technologies, {NAACL-HLT} 2019, Minneapolis, MN, USA, June 2-7,
  2019, Volume 1 (Long and Short Papers)}}. \bibinfo{publisher}{Association for
  Computational Linguistics}, \bibinfo{pages}{4171--4186}.
\newblock
\urldef\tempurl%
\url{https://doi.org/10.18653/v1/n19-1423}
\showDOI{\tempurl}


\bibitem[Dong et~al\mbox{.}(2014)]%
        {dongGHHLMSSZ14}
\bibfield{author}{\bibinfo{person}{Xin Dong}, \bibinfo{person}{Evgeniy
  Gabrilovich}, \bibinfo{person}{Geremy Heitz}, \bibinfo{person}{Wilko Horn},
  \bibinfo{person}{Ni Lao}, \bibinfo{person}{Kevin Murphy},
  \bibinfo{person}{Thomas Strohmann}, \bibinfo{person}{Shaohua Sun}, {and}
  \bibinfo{person}{Wei Zhang}.} \bibinfo{year}{2014}\natexlab{}.
\newblock \showarticletitle{Knowledge vault: a web-scale approach to
  probabilistic knowledge fusion}. In \bibinfo{booktitle}{\emph{The 20th {ACM}
  {SIGKDD} International Conference on Knowledge Discovery and Data Mining,
  {KDD} '14, New York, NY, {USA} - August 24 - 27, 2014}}.
  \bibinfo{publisher}{{ACM}}, \bibinfo{pages}{601--610}.
\newblock
\urldef\tempurl%
\url{https://doi.org/10.1145/2623330.2623623}
\showDOI{\tempurl}


\bibitem[Fahandar and H{\"{u}}llermeier(2018)]%
        {FahandarH18}
\bibfield{author}{\bibinfo{person}{Mohsen~Ahmadi Fahandar} {and}
  \bibinfo{person}{Eyke H{\"{u}}llermeier}.} \bibinfo{year}{2018}\natexlab{}.
\newblock \showarticletitle{Learning to Rank Based on Analogical Reasoning}. In
  \bibinfo{booktitle}{\emph{Proceedings of the Thirty-Second {AAAI} Conference
  on Artificial Intelligence, (AAAI-18), the 30th innovative Applications of
  Artificial Intelligence (IAAI-18), and the 8th {AAAI} Symposium on
  Educational Advances in Artificial Intelligence (EAAI-18), New Orleans,
  Louisiana, USA, February 2-7, 2018}}. \bibinfo{publisher}{{AAAI} Press},
  \bibinfo{pages}{2951--2958}.
\newblock


\bibitem[Fahandar and H{\"{u}}llermeier(2021)]%
        {FahandarH21}
\bibfield{author}{\bibinfo{person}{Mohsen~Ahmadi Fahandar} {and}
  \bibinfo{person}{Eyke H{\"{u}}llermeier}.} \bibinfo{year}{2021}\natexlab{}.
\newblock \showarticletitle{Analogical Embedding for Analogy-Based Learning to
  Rank}. In \bibinfo{booktitle}{\emph{Advances in Intelligent Data Analysis
  {XIX} - 19th International Symposium on Intelligent Data Analysis, {IDA}
  2021, Porto, Portugal, April 26-28, 2021, Proceedings}}
  \emph{(\bibinfo{series}{Lecture Notes in Computer Science},
  Vol.~\bibinfo{volume}{12695})}. \bibinfo{publisher}{Springer},
  \bibinfo{pages}{76--88}.
\newblock
\urldef\tempurl%
\url{https://doi.org/10.1007/978-3-030-74251-5_7}
\showDOI{\tempurl}


\bibitem[Faralli et~al\mbox{.}(2018)]%
        {faralli18}
\bibfield{author}{\bibinfo{person}{Stefano Faralli}, \bibinfo{person}{Irene
  Finocchi}, \bibinfo{person}{Simone~Paolo Ponzetto}, {and}
  \bibinfo{person}{Paola Velardi}.} \bibinfo{year}{2018}\natexlab{}.
\newblock \showarticletitle{Efficient Pruning of Large Knowledge Graphs}. In
  \bibinfo{booktitle}{\emph{Proceedings of the Twenty-Seventh International
  Joint Conference on Artificial Intelligence, {IJCAI} 2018, July 13-19, 2018,
  Stockholm, Sweden}}. \bibinfo{publisher}{ijcai.org},
  \bibinfo{pages}{4055--4063}.
\newblock


\bibitem[Fernández-López et~al\mbox{.}(1997)]%
        {fernandez1997}
\bibfield{author}{\bibinfo{person}{Mariano Fernández-López},
  \bibinfo{person}{Asuncion Gomez-Perez}, {and} \bibinfo{person}{Natalia
  Juristo}.} \bibinfo{year}{1997}\natexlab{}.
\newblock \showarticletitle{METHONTOLOGY: from ontological art towards
  ontological engineering}.
\newblock \bibinfo{journal}{\emph{Engineering Workshop on Ontological
  Engineering (AAAI97)}} (\bibinfo{date}{03} \bibinfo{year}{1997}).
\newblock


\bibitem[Hogan et~al\mbox{.}(2021)]%
        {hoganBCAMGKENNNPRRSSSZ21}
\bibfield{author}{\bibinfo{person}{Aidan Hogan}, \bibinfo{person}{Eva
  Blomqvist}, \bibinfo{person}{Michael Cochez}, \bibinfo{person}{Claudia
  d'Amato}, \bibinfo{person}{Gerard de Melo}, \bibinfo{person}{Claudio
  Gutierrez}, \bibinfo{person}{Sabrina Kirrane}, \bibinfo{person}{Jos{\'{e}}
  Emilio~Labra Gayo}, \bibinfo{person}{Roberto Navigli},
  \bibinfo{person}{Sebastian Neumaier}, \bibinfo{person}{Axel{-}Cyrille~Ngonga
  Ngomo}, \bibinfo{person}{Axel Polleres}, \bibinfo{person}{Sabbir~M. Rashid},
  \bibinfo{person}{Anisa Rula}, \bibinfo{person}{Lukas Schmelzeisen},
  \bibinfo{person}{Juan Sequeda}, \bibinfo{person}{Steffen Staab}, {and}
  \bibinfo{person}{Antoine Zimmermann}.} \bibinfo{year}{2021}\natexlab{}.
\newblock \bibinfo{booktitle}{\emph{Knowledge Graphs}}.
\newblock \bibinfo{publisher}{Morgan {\&} Claypool Publishers}.
\newblock
\urldef\tempurl%
\url{https://doi.org/10.2200/S01125ED1V01Y202109DSK022}
\showDOI{\tempurl}


\bibitem[Hug et~al\mbox{.}(2019)]%
        {HugPRS19}
\bibfield{author}{\bibinfo{person}{Nicolas Hug}, \bibinfo{person}{Henri Prade},
  \bibinfo{person}{Gilles Richard}, {and} \bibinfo{person}{Mathieu Serrurier}.}
  \bibinfo{year}{2019}\natexlab{}.
\newblock \showarticletitle{Analogical proportion-based methods for
  recommendation - First investigations}.
\newblock \bibinfo{journal}{\emph{Fuzzy Sets Systems}}  \bibinfo{volume}{366}
  (\bibinfo{year}{2019}), \bibinfo{pages}{110--132}.
\newblock
\urldef\tempurl%
\url{https://doi.org/10.1016/j.fss.2018.11.007}
\showDOI{\tempurl}


\bibitem[Ilievski et~al\mbox{.}(2022)]%
        {ilievskiPS22}
\bibfield{author}{\bibinfo{person}{Filip Ilievski}, \bibinfo{person}{Jay
  Pujara}, {and} \bibinfo{person}{Kartik Shenoy}.}
  \bibinfo{year}{2022}\natexlab{}.
\newblock \showarticletitle{Does Wikidata Support Analogical Reasoning?}. In
  \bibinfo{booktitle}{\emph{Knowledge Graphs and Semantic Web - 4th
  Iberoamerican Conference and third Indo-American Conference, {KGSWC} 2022,
  Madrid, Spain, November 21-23, 2022, Proceedings}}
  \emph{(\bibinfo{series}{Communications in Computer and Information Science},
  Vol.~\bibinfo{volume}{1686})}. \bibinfo{publisher}{Springer},
  \bibinfo{pages}{178--191}.
\newblock
\urldef\tempurl%
\url{https://doi.org/10.1007/978-3-031-21422-6_13}
\showDOI{\tempurl}


\bibitem[Jarnac and Monnin(2022)]%
        {jarnacM22}
\bibfield{author}{\bibinfo{person}{Lucas Jarnac} {and} \bibinfo{person}{Pierre
  Monnin}.} \bibinfo{year}{2022}\natexlab{}.
\newblock \showarticletitle{Wikidata to Bootstrap an Enterprise Knowledge
  Graph: How to Stay on Topic?}. In \bibinfo{booktitle}{\emph{Proceedings of
  the 3rd Wikidata Workshop 2022 co-located with the 21st International
  Semantic Web Conference (ISWC2022), Virtual Event, Hanghzou, China, October
  2022}} \emph{(\bibinfo{series}{{CEUR} Workshop Proceedings},
  Vol.~\bibinfo{volume}{3262})}. \bibinfo{publisher}{CEUR-WS.org}.
\newblock
\urldef\tempurl%
\url{https://ceur-ws.org/Vol-3262/paper16.pdf}
\showURL{%
\tempurl}


\bibitem[Ji et~al\mbox{.}(2022)]%
        {ji2022}
\bibfield{author}{\bibinfo{person}{Shaoxiong Ji}, \bibinfo{person}{Shirui Pan},
  \bibinfo{person}{Erik Cambria}, \bibinfo{person}{Pekka Marttinen}, {and}
  \bibinfo{person}{Philip~S. Yu}.} \bibinfo{year}{2022}\natexlab{}.
\newblock \showarticletitle{A Survey on Knowledge Graphs: Representation,
  Acquisition, and Applications}.
\newblock \bibinfo{journal}{\emph{{IEEE} Transactions on Neural Networks and
  Learning Systems}} \bibinfo{volume}{33}, \bibinfo{number}{2}
  (\bibinfo{year}{2022}), \bibinfo{pages}{494--514}.
\newblock
\urldef\tempurl%
\url{https://doi.org/10.1109/TNNLS.2021.3070843}
\showDOI{\tempurl}


\bibitem[Langlais et~al\mbox{.}(2009)]%
        {analogy-alea:2009:langlais}
\bibfield{author}{\bibinfo{person}{Philippe Langlais},
  \bibinfo{person}{Fran{\c{c}}ois Yvon}, {and} \bibinfo{person}{Pierre
  Zweigenbaum}.} \bibinfo{year}{2009}\natexlab{}.
\newblock \showarticletitle{Improvements in Analogical Learning: Application to
  Translating Multi-Terms of the Medical Domain}. In
  \bibinfo{booktitle}{\emph{{EACL} 2009, 12th Conference of the European
  Chapter of the Association for Computational Linguistics, Proceedings of the
  Conference, Athens, Greece, March 30 - April 3, 2009}}.
  \bibinfo{publisher}{The Association for Computer Linguistics},
  \bibinfo{pages}{487--495}.
\newblock
\urldef\tempurl%
\url{https://aclanthology.org/E09-1056/}
\showURL{%
\tempurl}


\bibitem[Lerer et~al\mbox{.}(2019)]%
        {lerer2019}
\bibfield{author}{\bibinfo{person}{Adam Lerer}, \bibinfo{person}{Ledell Wu},
  \bibinfo{person}{Jiajun Shen}, \bibinfo{person}{Timoth{\'{e}}e Lacroix},
  \bibinfo{person}{Luca Wehrstedt}, \bibinfo{person}{Abhijit Bose}, {and}
  \bibinfo{person}{Alex Peysakhovich}.} \bibinfo{year}{2019}\natexlab{}.
\newblock \showarticletitle{Pytorch-BigGraph: {A} Large Scale Graph Embedding
  System}. In \bibinfo{booktitle}{\emph{Proceedings of Machine Learning and
  Systems 2019, MLSys 2019, Stanford, CA, USA, March 31 - April 2, 2019}}.
  \bibinfo{publisher}{mlsys.org}.
\newblock


\bibitem[Lim et~al\mbox{.}(2019)]%
        {lim19}
\bibfield{author}{\bibinfo{person}{Suryani Lim}, \bibinfo{person}{Henri Prade},
  {and} \bibinfo{person}{Gilles Richard}.} \bibinfo{year}{2019}\natexlab{}.
\newblock \showarticletitle{Solving Word Analogies: {A} Machine Learning
  Perspective}. In \bibinfo{booktitle}{\emph{Symbolic and Quantitative
  Approaches to Reasoning with Uncertainty, 15th European Conference, {ECSQARU}
  2019, Belgrade, Serbia, September 18-20, 2019, Proceedings}}
  \emph{(\bibinfo{series}{Lecture Notes in Computer Science},
  Vol.~\bibinfo{volume}{11726})}. \bibinfo{publisher}{Springer},
  \bibinfo{pages}{238--250}.
\newblock
\urldef\tempurl%
\url{https://doi.org/10.1007/978-3-030-29765-7\_20}
\showDOI{\tempurl}


\bibitem[Lim et~al\mbox{.}(2021)]%
        {analogies-ml:2021:lim}
\bibfield{author}{\bibinfo{person}{Suryani Lim}, \bibinfo{person}{Henri Prade},
  {and} \bibinfo{person}{Gilles Richard}.} \bibinfo{year}{2021}\natexlab{}.
\newblock \showarticletitle{Classifying and completing word analogies by
  machine learning}.
\newblock \bibinfo{journal}{\emph{International Journal of Approximate
  Reasoning}}  \bibinfo{volume}{132} (\bibinfo{year}{2021}),
  \bibinfo{pages}{1--25}.
\newblock
\urldef\tempurl%
\url{https://doi.org/10.1016/j.ijar.2021.02.002}
\showDOI{\tempurl}


\bibitem[Liu et~al\mbox{.}(2017)]%
        {liuWY17}
\bibfield{author}{\bibinfo{person}{Hanxiao Liu}, \bibinfo{person}{Yuexin Wu},
  {and} \bibinfo{person}{Yiming Yang}.} \bibinfo{year}{2017}\natexlab{}.
\newblock \showarticletitle{Analogical Inference for Multi-relational
  Embeddings}. In \bibinfo{booktitle}{\emph{Proceedings of the 34th
  International Conference on Machine Learning, {ICML} 2017, Sydney, NSW,
  Australia, 6-11 August 2017}} \emph{(\bibinfo{series}{Proceedings of Machine
  Learning Research}, Vol.~\bibinfo{volume}{70})}. \bibinfo{publisher}{{PMLR}},
  \bibinfo{pages}{2168--2178}.
\newblock
\urldef\tempurl%
\url{http://proceedings.mlr.press/v70/liu17d.html}
\showURL{%
\tempurl}


\bibitem[Mahdisoltani et~al\mbox{.}(2015)]%
        {mahdisoltaniBS15}
\bibfield{author}{\bibinfo{person}{Farzaneh Mahdisoltani},
  \bibinfo{person}{Joanna Biega}, {and} \bibinfo{person}{Fabian~M. Suchanek}.}
  \bibinfo{year}{2015}\natexlab{}.
\newblock \showarticletitle{{YAGO3:} {A} Knowledge Base from Multilingual
  Wikipedias}. In \bibinfo{booktitle}{\emph{Seventh Biennial Conference on
  Innovative Data Systems Research, {CIDR} 2015, Asilomar, CA, USA, January
  4-7, 2015, Online Proceedings}}. \bibinfo{publisher}{www.cidrdb.org}.
\newblock
\urldef\tempurl%
\url{http://cidrdb.org/cidr2015/Papers/CIDR15\_Paper1.pdf}
\showURL{%
\tempurl}


\bibitem[Marquer et~al\mbox{.}(2022)]%
        {marquer22}
\bibfield{author}{\bibinfo{person}{Esteban Marquer}, \bibinfo{person}{Safa
  Alsaidi}, \bibinfo{person}{Amandine Decker},
  \bibinfo{person}{Pierre{-}Alexandre Murena}, {and} \bibinfo{person}{Miguel
  Couceiro}.} \bibinfo{year}{2022}\natexlab{}.
\newblock \showarticletitle{A Deep Learning Approach to Solving Morphological
  Analogies}. In \bibinfo{booktitle}{\emph{Case-Based Reasoning Research and
  Development - 30th International Conference, {ICCBR} 2022, Nancy, France,
  September 12-15, 2022, Proceedings}} \emph{(\bibinfo{series}{Lecture Notes in
  Computer Science}, Vol.~\bibinfo{volume}{13405})}.
  \bibinfo{publisher}{Springer}, \bibinfo{pages}{159--174}.
\newblock
\urldef\tempurl%
\url{https://doi.org/10.1007/978-3-031-14923-8\_11}
\showDOI{\tempurl}


\bibitem[Miclet et~al\mbox{.}(2008)]%
        {miclet2008}
\bibfield{author}{\bibinfo{person}{Laurent Miclet}, \bibinfo{person}{Sabri
  Bayoudh}, {and} \bibinfo{person}{Arnaud Delhay}.}
  \bibinfo{year}{2008}\natexlab{}.
\newblock \showarticletitle{Analogical Dissimilarity: Definition, Algorithms
  and Two Experiments in Machine Learning}.
\newblock \bibinfo{journal}{\emph{Journal of Artificial Intelligence Research}}
   \bibinfo{volume}{32} (\bibinfo{year}{2008}), \bibinfo{pages}{793--824}.
\newblock
\urldef\tempurl%
\url{https://doi.org/10.1613/jair.2519}
\showDOI{\tempurl}


\bibitem[Mikolov et~al\mbox{.}(2013a)]%
        {efficient-representation-w2v:2013:mikolov}
\bibfield{author}{\bibinfo{person}{Tom{\'{a}}s Mikolov}, \bibinfo{person}{Kai
  Chen}, \bibinfo{person}{Greg Corrado}, {and} \bibinfo{person}{Jeffrey Dean}.}
  \bibinfo{year}{2013}\natexlab{a}.
\newblock \showarticletitle{Efficient Estimation of Word Representations in
  Vector Space}. In \bibinfo{booktitle}{\emph{1st International Conference on
  Learning Representations, {ICLR} 2013, Scottsdale, Arizona, USA, May 2-4,
  2013, Workshop Track Proceedings}}.
\newblock
\urldef\tempurl%
\url{http://arxiv.org/abs/1301.3781}
\showURL{%
\tempurl}


\bibitem[Mikolov et~al\mbox{.}(2013b)]%
        {MikolovNIPS2013}
\bibfield{author}{\bibinfo{person}{Tom{\'{a}}s Mikolov}, \bibinfo{person}{Ilya
  Sutskever}, \bibinfo{person}{Kai Chen}, \bibinfo{person}{Gregory~S. Corrado},
  {and} \bibinfo{person}{Jeffrey Dean}.} \bibinfo{year}{2013}\natexlab{b}.
\newblock \showarticletitle{Distributed Representations of Words and Phrases
  and their Compositionality}.
\newblock In \bibinfo{booktitle}{\emph{Advances in Neural Information
  Processing Systems 26: 27th Annual Conference on Neural Information
  Processing Systems 2013. Proceedings of a meeting held December 5-8, 2013,
  Lake Tahoe, Nevada, United States}}. \bibinfo{pages}{3111--3119}.
\newblock
\urldef\tempurl%
\url{https://proceedings.neurips.cc/paper/2013/hash/9aa42b31882ec039965f3c4923ce901b-Abstract.html}
\showURL{%
\tempurl}


\bibitem[Mitchell(2021)]%
        {mitchell21}
\bibfield{author}{\bibinfo{person}{Melanie Mitchell}.}
  \bibinfo{year}{2021}\natexlab{}.
\newblock \showarticletitle{Abstraction and analogy-making in artificial
  intelligence}.
\newblock \bibinfo{journal}{\emph{Annals of the New York Academy of Sciences}}
  \bibinfo{volume}{1505}, \bibinfo{number}{1} (\bibinfo{year}{2021}),
  \bibinfo{pages}{79--101}.
\newblock


\bibitem[Monnin and Couceiro(2022)]%
        {monninC2022}
\bibfield{author}{\bibinfo{person}{Pierre Monnin} {and} \bibinfo{person}{Miguel
  Couceiro}.} \bibinfo{year}{2022}\natexlab{}.
\newblock \showarticletitle{Interactions Between Knowledge Graph-Related Tasks
  and Analogical Reasoning: {A} Discussion}. In
  \bibinfo{booktitle}{\emph{Workshop Proceedings of the 30th International
  Conferece on Case-Based Reasoning co-located with the 30th International
  Conference on Case-Based Reasoning {(ICCBR} 2022), Nancy (France), September
  12-15th, 2022}} \emph{(\bibinfo{series}{{CEUR} Workshop Proceedings},
  Vol.~\bibinfo{volume}{3389})}. \bibinfo{publisher}{CEUR-WS.org},
  \bibinfo{pages}{57--67}.
\newblock
\urldef\tempurl%
\url{https://ceur-ws.org/Vol-3389/ICCBR\_2022\_Workshop\_paper\_75.pdf}
\showURL{%
\tempurl}


\bibitem[Monnin et~al\mbox{.}(2019)]%
        {monninLHRTJNC19}
\bibfield{author}{\bibinfo{person}{Pierre Monnin}, \bibinfo{person}{Jo{\"{e}}l
  Legrand}, \bibinfo{person}{Graziella Husson}, \bibinfo{person}{Patrice
  Ringot}, \bibinfo{person}{Andon Tchechmedjiev},
  \bibinfo{person}{Cl{\'{e}}ment Jonquet}, \bibinfo{person}{Amedeo Napoli},
  {and} \bibinfo{person}{Adrien Coulet}.} \bibinfo{year}{2019}\natexlab{}.
\newblock \showarticletitle{PGxO and PGxLOD: a reconciliation of
  pharmacogenomic knowledge of various provenances, enabling further
  comparison}.
\newblock \bibinfo{journal}{\emph{{BMC} Bioinformatics}}
  \bibinfo{volume}{20-S}, \bibinfo{number}{4} (\bibinfo{year}{2019}),
  \bibinfo{pages}{139:1--139:16}.
\newblock
\urldef\tempurl%
\url{https://doi.org/10.1186/s12859-019-2693-9}
\showDOI{\tempurl}


\bibitem[Murena et~al\mbox{.}(2020)]%
        {minimal-complexity:2020:murena}
\bibfield{author}{\bibinfo{person}{Pierre{-}Alexandre Murena},
  \bibinfo{person}{Marie Al{-}Ghossein}, \bibinfo{person}{Jean{-}Louis
  Dessalles}, {and} \bibinfo{person}{Antoine Cornu{\'{e}}jols}.}
  \bibinfo{year}{2020}\natexlab{}.
\newblock \showarticletitle{Solving Analogies on Words based on Minimal
  Complexity Transformation}. In \bibinfo{booktitle}{\emph{Proceedings of the
  Twenty-Ninth International Joint Conference on Artificial Intelligence,
  {IJCAI} 2020}}. \bibinfo{publisher}{ijcai.org}, \bibinfo{pages}{1848--1854}.
\newblock
\urldef\tempurl%
\url{https://doi.org/10.24963/ijcai.2020/256}
\showDOI{\tempurl}


\bibitem[Noy et~al\mbox{.}(2019)]%
        {noyGJNPT19}
\bibfield{author}{\bibinfo{person}{Natalya~Fridman Noy},
  \bibinfo{person}{Yuqing Gao}, \bibinfo{person}{Anshu Jain},
  \bibinfo{person}{Anant Narayanan}, \bibinfo{person}{Alan Patterson}, {and}
  \bibinfo{person}{Jamie Taylor}.} \bibinfo{year}{2019}\natexlab{}.
\newblock \showarticletitle{Industry-scale knowledge graphs: lessons and
  challenges}.
\newblock \bibinfo{journal}{\emph{Commun. ACM}} \bibinfo{volume}{62},
  \bibinfo{number}{8} (\bibinfo{year}{2019}), \bibinfo{pages}{36--43}.
\newblock
\urldef\tempurl%
\url{https://doi.org/10.1145/3331166}
\showDOI{\tempurl}


\bibitem[Pedregosa et~al\mbox{.}(2011)]%
        {scikit-learn}
\bibfield{author}{\bibinfo{person}{Fabian Pedregosa},
  \bibinfo{person}{Ga{\"{e}}l Varoquaux}, \bibinfo{person}{Alexandre Gramfort},
  \bibinfo{person}{Vincent Michel}, \bibinfo{person}{Bertrand Thirion},
  \bibinfo{person}{Olivier Grisel}, \bibinfo{person}{Mathieu Blondel},
  \bibinfo{person}{Peter Prettenhofer}, \bibinfo{person}{Ron Weiss},
  \bibinfo{person}{Vincent Dubourg}, \bibinfo{person}{Jake VanderPlas},
  \bibinfo{person}{Alexandre Passos}, \bibinfo{person}{David Cournapeau},
  \bibinfo{person}{Matthieu Brucher}, \bibinfo{person}{Matthieu Perrot}, {and}
  \bibinfo{person}{Edouard Duchesnay}.} \bibinfo{year}{2011}\natexlab{}.
\newblock \showarticletitle{Scikit-learn: Machine Learning in {P}ython}.
\newblock \bibinfo{journal}{\emph{Journal of Machine Learning Research}}
  \bibinfo{volume}{12} (\bibinfo{year}{2011}), \bibinfo{pages}{2825--2830}.
\newblock


\bibitem[Peyre et~al\mbox{.}(2019)]%
        {PeyreSLS19}
\bibfield{author}{\bibinfo{person}{Julia Peyre}, \bibinfo{person}{Josef Sivic},
  \bibinfo{person}{Ivan Laptev}, {and} \bibinfo{person}{Cordelia Schmid}.}
  \bibinfo{year}{2019}\natexlab{}.
\newblock \showarticletitle{Detecting Unseen Visual Relations Using Analogies}.
  In \bibinfo{booktitle}{\emph{2019 {IEEE/CVF} International Conference on
  Computer Vision, {ICCV} 2019, Seoul, Korea (South), October 27 - November 2,
  2019}}. \bibinfo{publisher}{{IEEE}}, \bibinfo{pages}{1981--1990}.
\newblock
\urldef\tempurl%
\url{https://doi.org/10.1109/ICCV.2019.00207}
\showDOI{\tempurl}


\bibitem[Portisch et~al\mbox{.}(2022)]%
        {portischHP22}
\bibfield{author}{\bibinfo{person}{Jan Portisch}, \bibinfo{person}{Nicolas
  Heist}, {and} \bibinfo{person}{Heiko Paulheim}.}
  \bibinfo{year}{2022}\natexlab{}.
\newblock \showarticletitle{Knowledge graph embedding for data mining vs.
  knowledge graph embedding for link prediction - two sides of the same coin?}
\newblock \bibinfo{journal}{\emph{Semantic Web}} \bibinfo{volume}{13},
  \bibinfo{number}{3} (\bibinfo{year}{2022}), \bibinfo{pages}{399--422}.
\newblock
\urldef\tempurl%
\url{https://doi.org/10.3233/SW-212892}
\showDOI{\tempurl}


\bibitem[Sadeghi et~al\mbox{.}(2015)]%
        {SadeghiZF15}
\bibfield{author}{\bibinfo{person}{Fereshteh Sadeghi},
  \bibinfo{person}{C.~Lawrence Zitnick}, {and} \bibinfo{person}{Ali Farhadi}.}
  \bibinfo{year}{2015}\natexlab{}.
\newblock \showarticletitle{Visalogy: Answering Visual Analogy Questions}. In
  \bibinfo{booktitle}{\emph{Advances in Neural Information Processing Systems
  28: Annual Conference on Neural Information Processing Systems 2015, December
  7-12, 2015, Montreal, Quebec, Canada}}. \bibinfo{pages}{1882--1890}.
\newblock


\bibitem[Sequeda and Lassila(2021)]%
        {sequedaL2021}
\bibfield{author}{\bibinfo{person}{Juan Sequeda} {and} \bibinfo{person}{Ora
  Lassila}.} \bibinfo{year}{2021}\natexlab{}.
\newblock \bibinfo{booktitle}{\emph{Designing and Building Enterprise Knowledge
  Graphs}}.
\newblock \bibinfo{publisher}{Morgan {\&} Claypool Publishers}.
\newblock
\urldef\tempurl%
\url{https://doi.org/10.2200/S01105ED1V01Y202105DSK020}
\showDOI{\tempurl}


\bibitem[Shbita et~al\mbox{.}(2023)]%
        {shbitaGLDR23}
\bibfield{author}{\bibinfo{person}{Basel Shbita}, \bibinfo{person}{Anna~Lisa
  Gentile}, \bibinfo{person}{Pengyuan Li}, \bibinfo{person}{Chad DeLuca}, {and}
  \bibinfo{person}{Guang{-}Jie Ren}.} \bibinfo{year}{2023}\natexlab{}.
\newblock \showarticletitle{Understanding Customer Requirements - An Enterprise
  Knowledge Graph Approach}. In \bibinfo{booktitle}{\emph{The Semantic Web -
  20th International Conference, {ESWC} 2023, Hersonissos, Crete, Greece, May
  28 - June 1, 2023, Proceedings}} \emph{(\bibinfo{series}{Lecture Notes in
  Computer Science}, Vol.~\bibinfo{volume}{13870})}.
  \bibinfo{publisher}{Springer}, \bibinfo{pages}{625--643}.
\newblock
\urldef\tempurl%
\url{https://doi.org/10.1007/978-3-031-33455-9_37}
\showDOI{\tempurl}


\bibitem[Shenoy et~al\mbox{.}(2022)]%
        {shenoy22}
\bibfield{author}{\bibinfo{person}{Kartik Shenoy}, \bibinfo{person}{Filip
  Ilievski}, \bibinfo{person}{Daniel Garijo}, \bibinfo{person}{Daniel Schwabe},
  {and} \bibinfo{person}{Pedro~A. Szekely}.} \bibinfo{year}{2022}\natexlab{}.
\newblock \showarticletitle{A study of the quality of {W}ikidata}.
\newblock \bibinfo{journal}{\emph{Journal of Web Semantics}}
  \bibinfo{volume}{72} (\bibinfo{year}{2022}), \bibinfo{pages}{100679}.
\newblock


\bibitem[Sultan and Shahaf(2022)]%
        {sultanS22}
\bibfield{author}{\bibinfo{person}{Oren Sultan} {and} \bibinfo{person}{Dafna
  Shahaf}.} \bibinfo{year}{2022}\natexlab{}.
\newblock \showarticletitle{Life is a Circus and We are the Clowns:
  Automatically Finding Analogies between Situations and Processes}. In
  \bibinfo{booktitle}{\emph{Proceedings of the 2022 Conference on Empirical
  Methods in Natural Language Processing, {EMNLP} 2022, Abu Dhabi, United Arab
  Emirates, December 7-11, 2022}}. \bibinfo{publisher}{Association for
  Computational Linguistics}, \bibinfo{pages}{3547--3562}.
\newblock
\urldef\tempurl%
\url{https://aclanthology.org/2022.emnlp-main.232}
\showURL{%
\tempurl}


\bibitem[Swartout et~al\mbox{.}(1996)]%
        {swartoutPKR1996}
\bibfield{author}{\bibinfo{person}{Bill Swartout}, \bibinfo{person}{Ramesh
  Patil}, \bibinfo{person}{Kevin Knight}, {and} \bibinfo{person}{Tom Russ}.}
  \bibinfo{year}{1996}\natexlab{}.
\newblock \showarticletitle{Toward distributed use of large-scale ontologies}.
  In \bibinfo{booktitle}{\emph{Proceedings of the Tenth Workshop on Knowledge
  Acquisition for Knowledge-Based Systems}}, Vol.~\bibinfo{volume}{138}.
  \bibinfo{pages}{25}.
\newblock


\bibitem[Tiddi and Schlobach(2022)]%
        {tiddiS22}
\bibfield{author}{\bibinfo{person}{Ilaria Tiddi} {and} \bibinfo{person}{Stefan
  Schlobach}.} \bibinfo{year}{2022}\natexlab{}.
\newblock \showarticletitle{Knowledge graphs as tools for explainable machine
  learning: {A} survey}.
\newblock \bibinfo{journal}{\emph{Artificial Intelligence}}
  \bibinfo{volume}{302} (\bibinfo{year}{2022}), \bibinfo{pages}{103627}.
\newblock
\urldef\tempurl%
\url{https://doi.org/10.1016/j.artint.2021.103627}
\showDOI{\tempurl}


\bibitem[Turney(2008)]%
        {Turney.08}
\bibfield{author}{\bibinfo{person}{Peter~D. Turney}.}
  \bibinfo{year}{2008}\natexlab{}.
\newblock \showarticletitle{The Latent Relation Mapping Engine: Algorithm and
  Experiments}.
\newblock   \bibinfo{volume}{33} (\bibinfo{year}{2008}),
  \bibinfo{pages}{615--655}.
\newblock
\urldef\tempurl%
\url{https://doi.org/10.1613/jair.2693}
\showDOI{\tempurl}


\bibitem[Vrandecic and Kr{\"{o}}tzsch(2014)]%
        {vrandecic}
\bibfield{author}{\bibinfo{person}{Denny Vrandecic} {and}
  \bibinfo{person}{Markus Kr{\"{o}}tzsch}.} \bibinfo{year}{2014}\natexlab{}.
\newblock \showarticletitle{Wikidata: a free collaborative knowledgebase}.
\newblock \bibinfo{journal}{\emph{Commun. ACM}} \bibinfo{volume}{57},
  \bibinfo{number}{10} (\bibinfo{year}{2014}), \bibinfo{pages}{78--85}.
\newblock


\bibitem[Weikum et~al\mbox{.}(2021)]%
        {weikumDRS21}
\bibfield{author}{\bibinfo{person}{Gerhard Weikum}, \bibinfo{person}{Xin~Luna
  Dong}, \bibinfo{person}{Simon Razniewski}, {and} \bibinfo{person}{Fabian~M.
  Suchanek}.} \bibinfo{year}{2021}\natexlab{}.
\newblock \showarticletitle{Machine Knowledge: Creation and Curation of
  Comprehensive Knowledge Bases}.
\newblock \bibinfo{journal}{\emph{Foundations and Trends Databases}}
  \bibinfo{volume}{10}, \bibinfo{number}{2-4} (\bibinfo{year}{2021}),
  \bibinfo{pages}{108--490}.
\newblock


\bibitem[Yao et~al\mbox{.}(2023)]%
        {yaoZCHZC2023}
\bibfield{author}{\bibinfo{person}{Zhen Yao}, \bibinfo{person}{Wen Zhang},
  \bibinfo{person}{Mingyang Chen}, \bibinfo{person}{Yufeng Huang},
  \bibinfo{person}{Yi Yang}, {and} \bibinfo{person}{Huajun Chen}.}
  \bibinfo{year}{2023}\natexlab{}.
\newblock \showarticletitle{Analogical Inference Enhanced Knowledge Graph
  Embedding}. In \bibinfo{booktitle}{\emph{Thirty-Seventh {AAAI} Conference on
  Artificial Intelligence, {AAAI} 2023, Thirty-Fifth Conference on Innovative
  Applications of Artificial Intelligence, {IAAI} 2023, Thirteenth Symposium on
  Educational Advances in Artificial Intelligence, {EAAI} 2023, Washington, DC,
  USA, February 7-14, 2023}}. \bibinfo{publisher}{{AAAI} Press},
  \bibinfo{pages}{4801--4808}.
\newblock
\urldef\tempurl%
\url{https://ojs.aaai.org/index.php/AAAI/article/view/25605}
\showURL{%
\tempurl}


\bibitem[Zervakis et~al\mbox{.}(2022)]%
        {Zervakis}
\bibfield{author}{\bibinfo{person}{Georgios Zervakis},
  \bibinfo{person}{Emmanuel Vincent}, \bibinfo{person}{Miguel Couceiro},
  \bibinfo{person}{Marc Schoenauer}, {and} \bibinfo{person}{Esteban Marquer}.}
  \bibinfo{year}{2022}\natexlab{}.
\newblock \showarticletitle{An Analogy based Approach for Solving Target Sense
  Verification}. In \bibinfo{booktitle}{\emph{Proceedings of the 2022 6th
  International Conference on Natural Language Processing and Information
  Retrieval, {NLPIR} 2022, Bangkok, Thailand, December 16-18, 2022}}.
  \bibinfo{publisher}{{ACM}}, \bibinfo{pages}{144--151}.
\newblock
\urldef\tempurl%
\url{https://doi.org/10.1145/3582768.3582794}
\showDOI{\tempurl}


\bibitem[Zhu and de~Melo(2020)]%
        {zhu-de-melo-2020-sentence}
\bibfield{author}{\bibinfo{person}{Xunjie Zhu} {and} \bibinfo{person}{Gerard de
  Melo}.} \bibinfo{year}{2020}\natexlab{}.
\newblock \showarticletitle{Sentence Analogies: Linguistic Regularities in
  Sentence Embeddings}. In \bibinfo{booktitle}{\emph{Proceedings of the 28th
  International Conference on Computational Linguistics, {COLING} 2020,
  Barcelona, Spain (Online), December 8-13, 2020}}.
  \bibinfo{publisher}{International Committee on Computational Linguistics},
  \bibinfo{pages}{3389--3400}.
\newblock
\urldef\tempurl%
\url{https://doi.org/10.18653/v1/2020.coling-main.300}
\showDOI{\tempurl}


\end{thebibliography}

\end{document}